\documentclass[review]{elsarticle}
\usepackage[utf8]{inputenc}
\usepackage{hyperref}
\usepackage{graphicx}
\usepackage[caption=false]{subfig}
\usepackage{amssymb,amsfonts}
\usepackage{comment}
\usepackage{color}
\usepackage{url}
\usepackage{multirow}
\usepackage{siunitx}
\usepackage[caption=false]{subfig}

\newcommand{\method}{HOb2sRNN}

\begin{document}
\begin{frontmatter}

\title{Object-based multi-temporal and multi-source land cover mapping leveraging hierarchical class relationships}

\author{Yawogan Jean Eudes Gbodjo\corref{mycorrespondingauthor}}
\cortext[mycorrespondingauthor]{Corresponding author}
\ead{jean-eudes.gbodjo@irstea.fr}
\address{IRSTEA, UMR TETIS, University of Montpellier, France}

\author{Dino Ienco}
\ead{dino.ienco@irstea.fr}
\address{IRSTEA, UMR TETIS, LIRMM, University of Montpellier, France}

\author{Louise Leroux}
\ead{louise.leroux@cirad.fr}
\address{CIRAD, UPR A\"IDA, Dakar, S\'{e}n\'{e}gal}

\author{Roberto Interdonato}
\ead{roberto.interdonato@cirad.fr}
\address{CIRAD, UMR TETIS, Montpellier, France}

\author{Raffaele Gaetano}
\ead{raffaele.gaetano@cirad}
\address{CIRAD, UMR TETIS, Montpellier, France}

\author{Babacar Ndao}
\ead{babacar.ndao@cse.sn}
\address{CSE, University Cheikh Anta Diop, Dakar, S\'{e}n\'{e}gal}

\author{Stephane Dupuy}
\ead{stephane.dupuy@cirad.fr}
\address{CIRAD, UMR TETIS, Montpellier, France}

\begin{abstract}
\label{sec:abstract}
European satellite missions Sentinel-1 (S1) and Sentinel-2 (S2) provide at high spatial resolution and high revisit time, respectively, radar and optical images that support a wide range of Earth surface monitoring tasks such as Land Use/Land Cover mapping. A  long-standing  challenge  in  the  remote  sensing  community is about  how to efficiently exploit multiple sources of information and leverage their complementary. In this particular case, get the most out of radar and optical satellite image time series (SITS). Here, we propose to deal with land cover mapping through a deep learning framework especially tailored to leverage the multi-source complementarity provided by radar and optical SITS. The proposed architecture is based on an extension of Recurrent Neural Network (RNN) enriched via a customized attention mechanism capable to fit the specificity of SITS data. In addition, we propose a new pretraining strategy that exploits domain expert knowledge to guide the model parameter initialization. Thorough experimental  evaluations involving several machine learning competitors, on two contrasted study sites, have demonstrated the suitability of our new attention mechanism combined with the extend RNN model as well as the benefit/limit to inject domain expert knowledge in the neural network training process.
\end{abstract}


\end{frontmatter}

\section{Introduction} \label{sec:intro}
Remotely sensed data collected by modern Earth Observation systems such as the European Sentinel programme~\cite{Berger2012} are getting more and more consideration in last years to cope with Earth surface monitoring. In particular, the Sentinel-1 and Sentinel-2 missions are of interest, since they provide publicly available multi-temporal radar and optical images respectively, with high spatial resolution (up to 10 meters) and high revisit time (up to 5 days). Thanks to these unprecedented spatial and temporal resolutions, images coming from such sensors, can be arranged in Satellite Image Time Series (SITS). SITS have been employed to deal with several tasks in multiple domains ranging from ecology~\cite{KoleckaGPPV18}, agriculture~\cite{KussulLSS17}, land management planning~\cite{IngladaVATMR17}, forest and natural habitat monitoring~\cite{guttler2017,KhialiIT18}.

Among these fields, Land Use/Land cover (LULC) mapping is getting more and more attention in these last years~\cite{STEINHAUSEN2018595,Dinh18,Gbodjo2019,INTERDONATO201991} since it provides essential components on which further indicators can be built on~\cite{Mousavi2019}.
As example, an accurate mapping of croplands and crop type is the cornerstone of agricultural monitoring systems as they allows to provide information on crop production and hence on food security for developing countries or food production for global market.However, the mapping of croplands has been identified as an important gap in agricultural monitoring systems~\cite{FRITZ2019258}.

As regards LULC mapping, both radar and optical sources have been employed, often solely, disregarding the well-known complementary existing between them, as underlined by recent works~\cite{GaoMSH06, IENCO201911, IannelliG18}. Also, when both sources of information are jointly used, they are independently processed without really leveraging the interplay between them, as well as the spatial and temporal dependencies they carry out~\cite{SharmaHT18,Fernandez-Beltran18,Erinjery2018345,STEINHAUSEN2018595,TrichtGGP18}.

Furthermore, concerning LULC mapping domain, specific knowledge about LULC classes can be available. LULC classes can be categorized in a hierarchical representation where LULC types are organized via class/subclass relationships. For instance, agricultural land cover types can be organized in crop types and subsequently crop types in specific crops. A notable example is the Food and Agriculture Organization (FAO)--Land Cover Classification System (LCCS)~\cite{di2005land} which is based on hierarchical and pre-defined approach in order to fit needs of any region of the world. 
Due to the presence of such class/subclass relationships, we can derive a hierarchical or taxonomic organization of LULC classes which could be appealing to consider in subsequent land mapping process. Only few studies, today, have considered the use of such hierarchical information to deal with land cover mapping~\cite{SULLAMENASHE2011392,Wu2016,SULLAMENASHE2019183}. Generally, such frameworks build an independent classification model for each level of the hierarchy and the decision made at a certain level of the taxonomy cannot be modified, further, in the decision process.

When dealing with land cover mapping, another challenge to deal with is related to the spatial granularity at which the remote sensing time series data is analyzed: pixel or object~\cite{blaschke2010}. While in the pixel based analysis, the basic units is the pixel, in object-based analysis, the images are first segmented and these segments (objects) become the basic units in any further analysis. Considering objects instead of pixels has two main advantages: i) objects represent a more coherent piece of information since they are simpler to interpret~\cite{lillesand2015} and ii) objects facilitate data analysis scale-up since, for the same image, the number of objects is usually smaller than the number of pixels by several order of magnitude. In addition, while extensive studies already exist on multi-temporal/multi-source land cover mapping at pixel level~\cite{IENCO201911, KussulLSS17}, no extensive evaluations are reported in literature about the possibility to combine multi-source data at object level.

Nowadays, Deep Learning (DL) is pervasive in many domains including remote sensing~\cite{Zhu17,IencoGDM17,MouGZ18,Zhong2019}.
Considering multi-source (radar and optical) data for LULC mapping, \cite{KussulLSS17} used a CNN-based architecture to combine Sentinel-1 and Landsat-8 images for land cover and crop types mapping. The CNN architecture processed the data with convolutions in spatial and spectral domains while the temporal domain was not take into account. Recently, \cite{IENCO201911} proposes TWINNS architecture a combination of CNN and RNN aiming to leverage both spatial and temporal dependencies in the SITS as well as the complementarity of radar and optical SITS. As underlined before, such method work at pixel level while they are not directly transferable to object-level analysis.

In this work, we propose to deal with land cover mapping, at object level, through a deep learning framework especially tailored to leverage both multi-source complementarity and temporal dependencies carried out by radar and optical SITS. The proposed architecture, named \method{} (Hierarchical Object based two-Stream Recurrent Neural Network), is based on an extension of Recurrent Neural Network (RNN) enriched via a customized attention mechanism capable to fit the specificity of SITS data. In addition, a new hierarchical pretraining strategy is proposed to exploit domain expert knowledge to guide the model parameter initialization. Thorough experimental evaluations were conducted on two study sites characterized by diverse land cover characteristics i.e. the Reunion island and a part of the Senegalese groundnut basin.

\section{Data and preprocessing} \label{sec:data}
The analysis was carried out on 2 study sites characterized by different landscapes and land cover classes : the Reunion island, a french overseas department located in the Indian Ocean and the southern part of the Senegalese groundnut basin located in the west center of Senegal. The Reunion island covers a little over 3000~$km^{2}$ of total area while the Senegalese site area is about 500~$km^{2}$. The former (resp. the latter) benchmark consists of 26 (resp. 16) Sentinel-1 (S1) images in radar wavelengths and 21 (resp. 19) Sentinel-2 (S2) images in optical wavelengths acquired between January and December 2017 (resp. May and October 2018).

\subsection{Sentinel-1 Data}
The radar data consists of S1 images acquired in C-band Interferometric Wide Swath (IW) mode with dual polarization (VH and VV) in ascending orbit. All images, as retrieved at level-1C Ground Range Detected (GRD) from the PEPS platform~\footnote{\url{https://peps.cnes.fr/}}, were first radiometrically calibrated in backscatter values (decibels, dB) using parameters included in the metadata file, then coregistered with the Sentinel-2 grid and orthorectified at the same 10-m spatial resolution. Finally, a multi-temporal filtering was applied to the time series removing artefacts resulting from speckle effect.

\subsection{Sentinel-2 Data}
The optical images were downloaded from the THEIA pole platform~\footnote{\url{http://theia.cnes.fr}} at level-2A top of canopy (TOC) reflectance. Only 10-m spatial resolution bands (i.e. Blue, Green, Red and Near Infrared spectrum) containing less than 50\% of cloudy pixels were considered in this analysis. The main issue with optical data, especially in tropical areas, is cloudiness. Therefore, a preprocessing was performed over each band to replace cloudy observations as detected by the supplied cloud masks through a multi-temporal gapfilling~\cite{IngladaVATMR17}. Cloudy pixel values were linearly interpolated using the previous and following cloud-free dates. The Normalized Difference Vegetation Index (NDVI)~\cite{Rouse1974} was calculated for each date. The NDVI was chosen as supplementary optical descriptor since it describes the photosynthetic activity and captures the metabolism intensity of the vegetation which is subject to changes in land cover.NDVI is thus considered as a reliable indicator to discriminate between different land cover classes and their changes over time.

\subsection{Ground truth} \label{subsec:GT}
Considering the Reunion island~\footnote{Reunion island learning database is available online on the CIRAD dataverse under doi:10.18167/DVN1/TOARDN}, the ground truth (GT) was built from various sources : the Registre Parcellaire Graphique (RPG)~\footnote{RPG is part of the European Land Parcel Identification  System (LPIS), provided by the French Agency for services and payment} reference data for 2014, GPS land cover records from June 2017 and the visual interpretation of very high spatial resolution (VHSR) SPOT6/7 images (1,5-m) completed by an expert with knowledge of territory to distinguish natural and urban areas. As regard the Senegalese site GT was built from GPS records collected during the 2018 agricultural campaign and the visual interpretation of a VHSR PlanetScope image (3-m). All GT were built in GIS vector file format containing a collection of polygons each attributed with the corresponding land cover class label. The Reunion island GT includes 6\,265 polygons distributed over 11 classes while the Senegalese site, which is less densely labeled like the former benchmark, includes 734 polygons distributed over 9 classes (See Tables~\ref{tab:gtreunion} and \ref{tab:gtsenegal}). 

\begin{table}[!htbp]
\caption{Characteristics of the Reunion island ground truth}
\label{tab:gtreunion}
\centering
 \begin{tabular}{|c|c|c|c|}
  \hline
  Class &  Label & Polygons & Segments \\
  \hline
  0 &  \emph{Sugar cane} & 869 & 1258 \\
  1 & \emph{Pasture and fodder} & 582 & 869 \\
  2 & \emph{Market gardening} & 758 & 912 \\
  3	& \emph{Greenhouse crops or shadows} & 260 & 233 \\
  4	& \emph{Orchards} & 767 & 1014 \\
  5	& \emph{Wooded areas} & 570 & 1106 \\
  6	& \emph{Moor and Savannah} & 506 & 850 \\
  7	& \emph{Rocks and natural bare soil} & 299 & 573 \\
  8	& \emph{Relief shadows} & 81 & 107 \\
  9 & \emph{Water} & 177 & 261 \\
 10	& \emph{Urbanized areas} & 1396 & 725 \\ \hline
 Total & & 6265 & 7908 \\
  \hline
 \end{tabular}
\end{table}

\begin{table}[!htbp]
\caption{Characteristics of the Senegalese site ground truth}
\label{tab:gtsenegal}
\centering
 \begin{tabular}{|c|c|c|c|}
  \hline
   Class &  Label &  Polygons &  Segments \\
  \hline
    0 &  \emph{Bushes} & 50 & 100 \\
    1 &  \emph{Fallows and Uncultivated areas} & 69 & 322 \\
    2 &  \emph{Ponds} & 33 & 59 \\
    3 &  \emph{Banks and bare soils} & 35 & 132 \\
    4 &  \emph{Villages} & 21 & 767 \\
    5 &  \emph{Wet areas} & 22 & 156 \\
    6 &  \emph{Valley} & 22 & 56 \\
    7 &  \emph{Cereals} & 260 & 816 \\
    8 &  \emph{Legumes} & 222 & 676 \\ \hline
    Total & & 734 & 3084 \\
  \hline
  \end{tabular}
\end{table}

\begin{figure}[!htbp]
\centering
\includegraphics[width=\columnwidth]{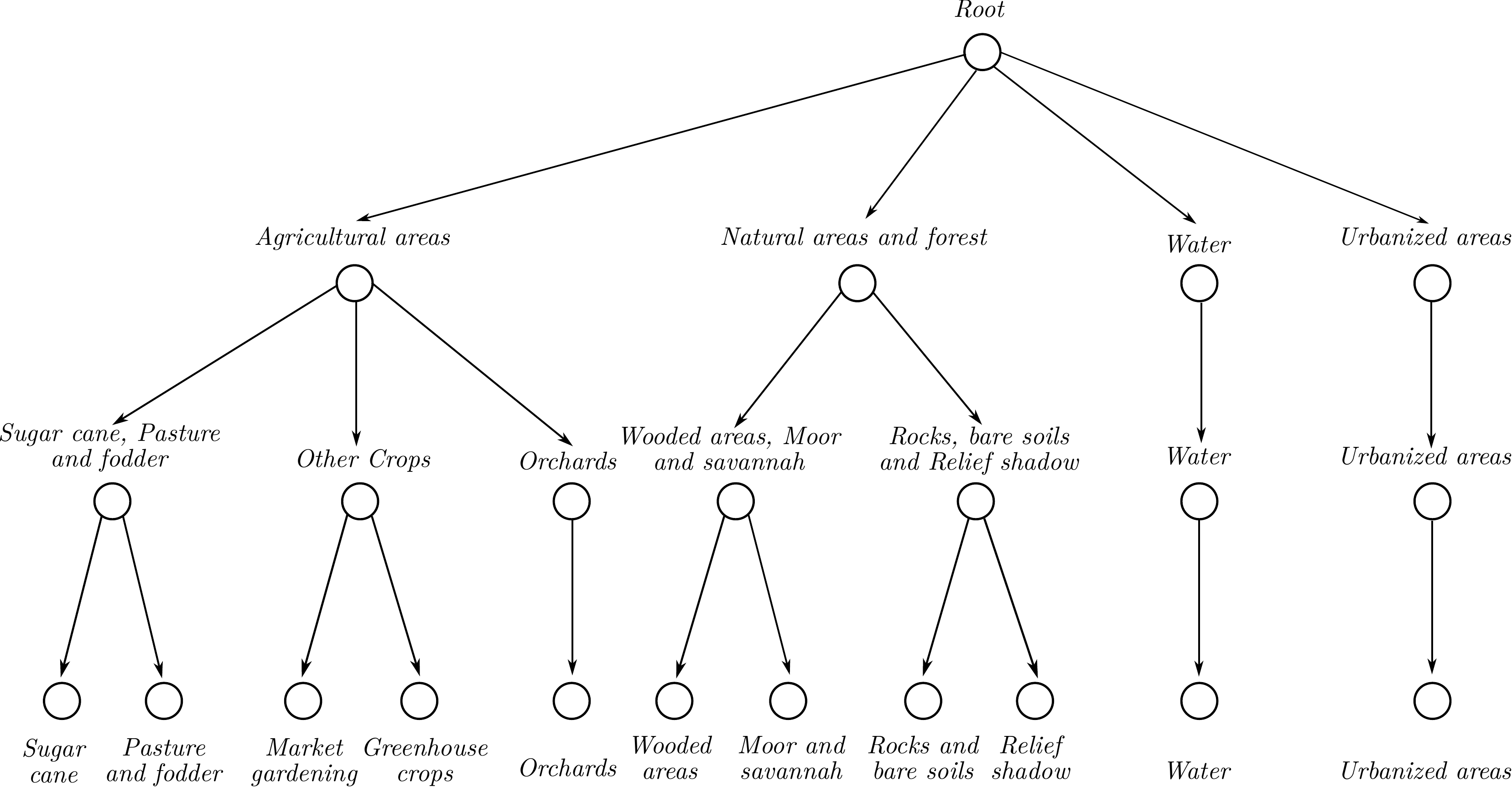}
\caption{ \label{fig:hier-reunion} Representation of the Reunion island hierarchical organization}
\end{figure}

\begin{figure}[!htbp]
\centering
\includegraphics[width=\columnwidth]{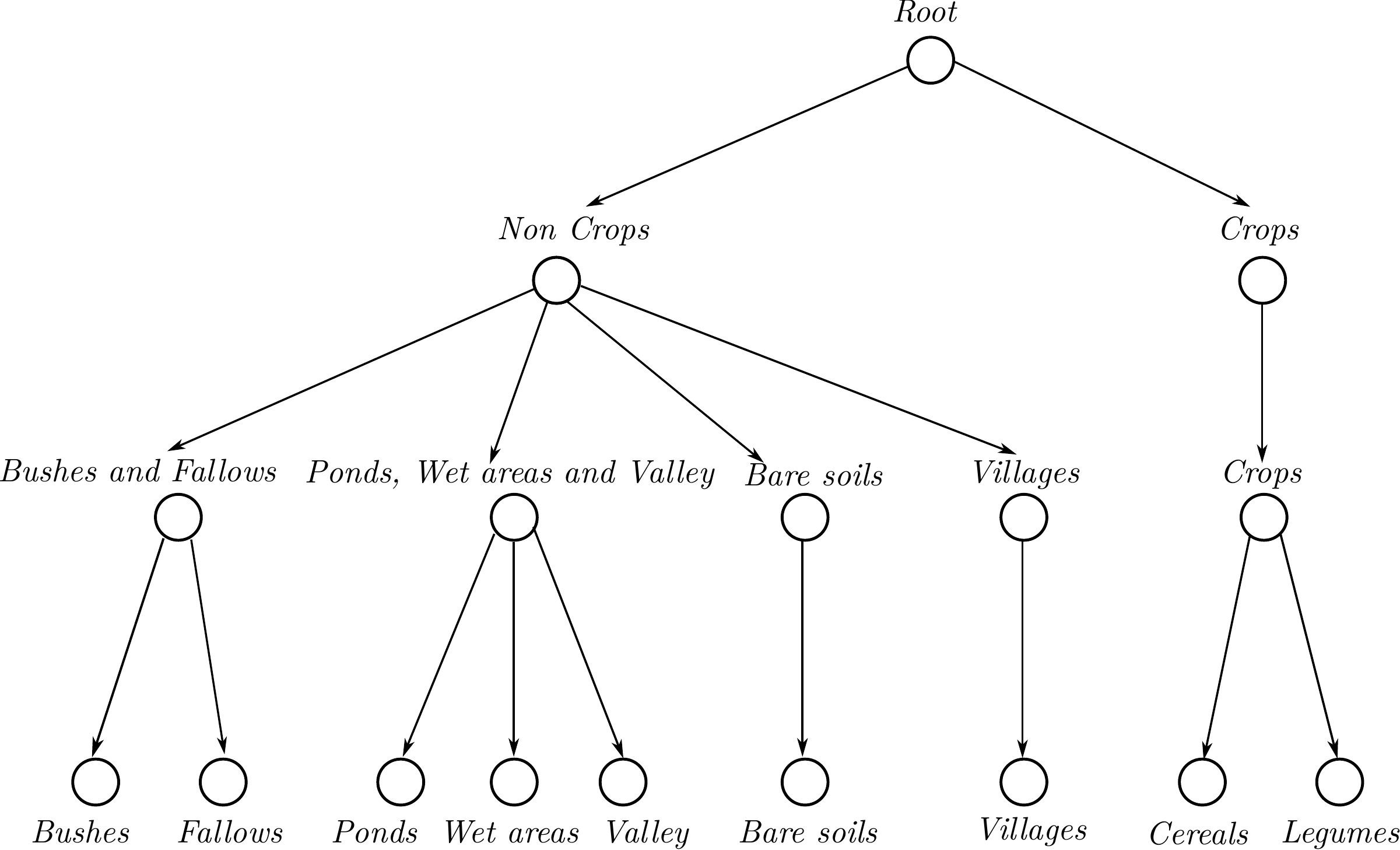}
\caption{ \label{fig:hier-senegal} Representation of the Senegalese site hierarchical organization}
\end{figure}

In order to integrate specific knowledge in the land cover mapping process, we derive for each study site a hierarchical organization of land cover classes (See Figures~\ref{fig:hier-reunion} and \ref{fig:hier-senegal}) obtaining two levels before the target classification level described in Tables~\ref{tab:gtreunion} and \ref{tab:gtsenegal}. 

To analyse data at object-level, a segmentation was performed for each study site using the VHSR images at hand (i.e. SPOT6/7 and PlanetScope) which have been coregistered with the corresponding Sentinel-2 grid to ensure a precise spatial matching. The VHSR images were segmented using the Large Scale Generic Region Merging (LSGRM) Orfeo Toolbox remote module~\cite{Lassalle15} obtaining 14\,465 (resp. 116\,937) segments for the Reunion island (resp. the Senegalese site). Segmentation parameters were ajusted so that the obtained segments fit as closely as possible field plot boundaries. Then, for each study site, the ground truth data were spatially intersected with the obtained segments to provide radiometrically homogeneous class samples and it finally resulted in new comparable size labeled 7\,908 segments for the Reunion island (resp. 3\,084 segments for the Senegalese site). See Tables~\ref{tab:gtreunion} and \ref{tab:gtsenegal} for details. Finally, the mean value of the pixels corresponding to each segment was calculated over all the timestamps in the SITS, resulting in 157 variables per segment ($26\times2$ for S1 $+$ $21\times5$ for S2) for Reunion and 127 variables per segment ($19\times5$ for S2 $+$ $16\times2$ for S1) for the Senegalese groundnut basin.

\section{Method} \label{sec:method}
\begin{figure}[!htbp]
\centering
\includegraphics[width=0.9\columnwidth]{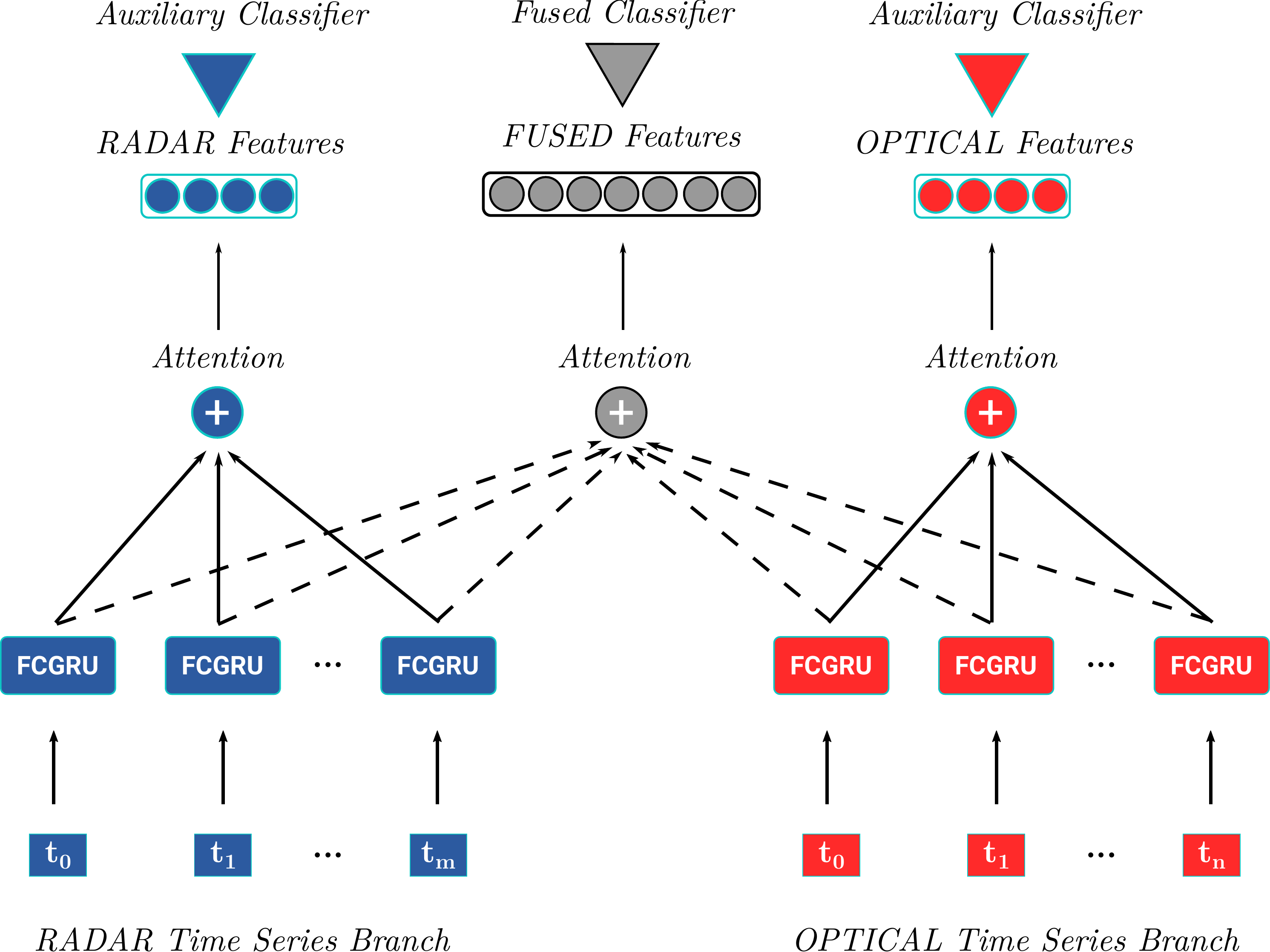}
\caption{ \label{fig:method}  \method{} takes as input two time series (radar and optical SITS) and provides as output the land cover classification. The architecture is composed of two branches. Each branch firstly processes the SITS by means of a modified Gated Recurrent Unit cell we named FCGRU and, successively, a modified attention mechanism is employed on top of the FCGRU outputs to extract the per source features. Furthermore, the same attention mechanism is employed on the concatenation of the per source outputs allowing to extract fused features. Finally, the set of features (per-source and fused) extracted is leveraged to provide the final classification.}
\end{figure}

Figure~\ref{fig:method} depicts the proposed deep learning architecture, named \method{}, for the multi-source SITS classification process. The architecture involves two branches: one for the radar (left) and one for the optical (right) time series. The output of the model is a land cover classification for each pair time series (radar and optical). Each branch of the \method{} architecture can be approximately decomposed in two parts: i) the time series analysis through the modified Gated Recurrent Unit cell we named FCGRU and ii) the multi-temporal combination to generate per-source features employing a modified attention mechanism. Moreover, the per branch FCGRU outputs are concatenated and the same attention mechanism is employed to extract fused features. Finally, the extracted per branch and fused features are leveraged to produce the final land cover classification. Such learned features named $feat_{rad}$, $feat_{opt}$ and $feat_{fused}$ indicate respectively the output of the radar branch, the optical branch and the \textcolor{black}{two} sources combination process. The architecture is trained leveraging specific knowledge derived from the hierarchical organization of the land cover classes. 

\subsection{Fully Connected Gated Recurrent Unit (FCGRU)}

The first part of each branch is constituted by a modified structure of Gated Recurrent Unit (GRU) cell we named Fully Connected GRU (FCGRU). GRU~\cite{ChoMGBBSB14} \textcolor{black}{is} a kind of Recurrent Neural Network (RNN) like Long Term Short Memory (LSTM)~\cite{HochreiterS96}, which has demonstrated its effectiveness in the field of remote sensing~\cite{Benedetti18,MouGZ17} among others. Unlike standard feed forward networks (i.e. CNNs), RNNs explicitly manage temporal (\textcolor{black}{sequential}) data dependencies since the output of the neuron at time $t-1$ is used, together with the next input, to feed the neuron itself at time $t$. Furthermore, this approach explicitly models the temporal correlation presents in the object time series and is able to focus its analysis on the useful portion of the time series (i.e., discarding less useful information). 

\begin{figure}[!htbp]
\centering
\includegraphics[width=\columnwidth]{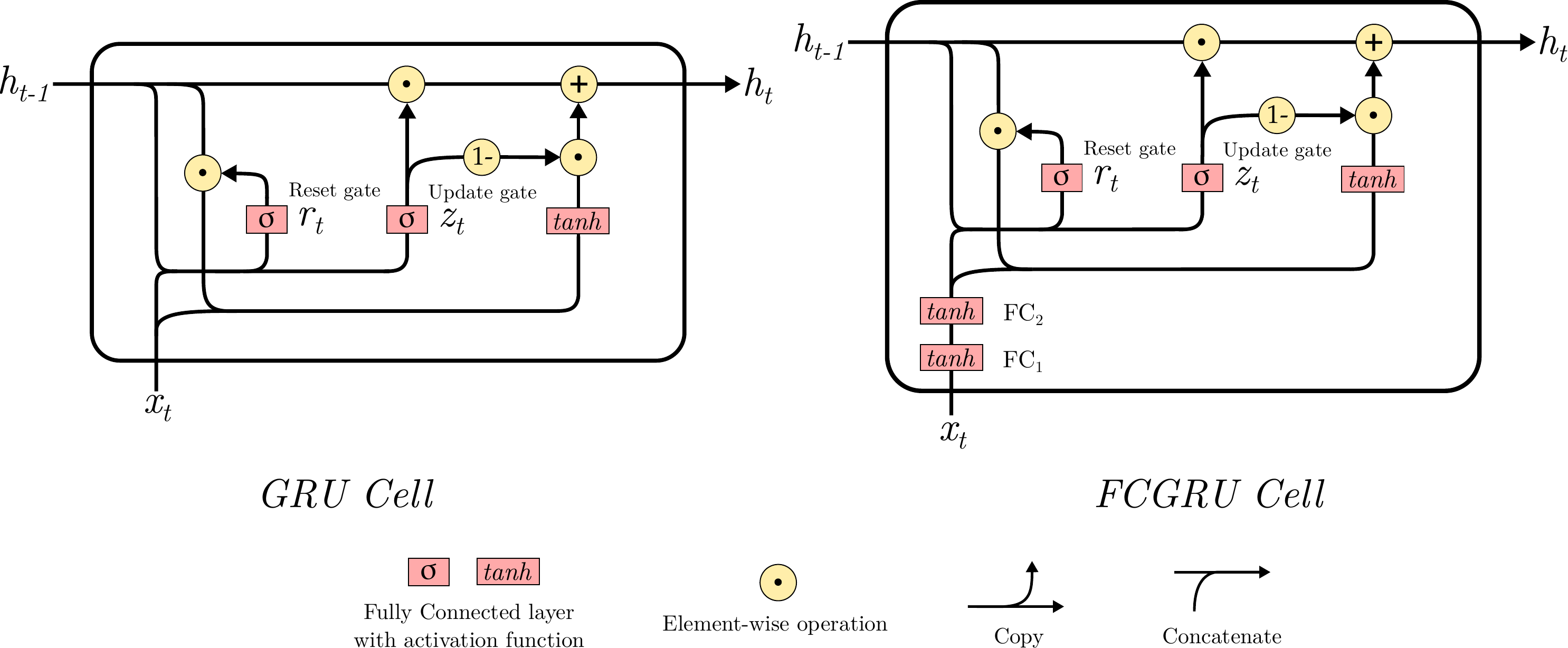}
\caption{ \label{fig:fcgru} Visual representation of the GRU and FCGRU cells }
\end{figure}

\textcolor{black}{In Figure~\ref{fig:fcgru} we compare the standard GRU unit with the introduced FCGRU. The main difference} 

is related \textcolor{black}{to the fact that the latter involves two} fully connected layers ($FC_1$ and $FC_2$) \textcolor{black}{to preprocess the time series information before start the standard transformation involved in the GRU unit.} 
This layer takes as input one sequence (i.e. time stamp) of the object time series (radar or optical) and combine the input data. Such layers allow the architecture to extract an useful input combination for the classification task enriching the original data representation. A Hyperbolic Tangent non-linearity function ($tanh$) is associated to each of the fully connected layers. \textcolor{black}{Successively, the standard GRU unit is employed. It is composed} 

of a hidden state $h_t-1$, and two different gates: the reset gate $r_t$ and the update gate $z_t$. The gates have two important functions: i) they regulate how much information has to be forgotten/remembered during the process and ii) they deal with the vanishing/exploding gradient problem. The gates are implemented by a Sigmoid function ($\sigma$) returning values between 0 and 1. The output of the unit is the new hidden state $h_t$. The following equations formally describe the FCGRU cell:
\begin{equation}
x_{t'} = \tanh (W_2 \tanh( W_1 x_t + b_1 ) + b_2) \label{eqn:gru1}
\end{equation}
\begin{equation}
z_{t} = \sigma(W_{zx} x_{t'}  + W_{zh} h_{t-1} + b_z  ) \label{eqn:gru2}
\end{equation}
\begin{equation}
r_{t} = \sigma(W_{rx} x_{t'}  + W_{rh} h_{t-1} + b_r  ) 
\label{eqn:gru3}
\end{equation}
\begin{equation}
h_{t} = z_t \odot h_{t-1} + (1-z_t) \odot \tanh( W_{hx} x_{t} + W_{hr} (r_t \odot h_{t-1})+b_h  )
\label{eqn:gru4}
\end{equation}

The $\odot$ symbol indicates an element-wise multiplication while $\sigma$ and $\tanh$ represent Sigmoid and Hyperbolic Tangent function, respectively.  $x_t$ is the time stamp input vector and $x_{t'}$ is the enriched input vector representation. The different $W_{*}$, $W_{**}$ matrices and bias coefficients $b_{*}$ are the parameters learned during the training of the model. Dropout was employed in the FCGRU cell and between the two fully connected layers to prevent overfitting. 

\subsection{Modified Attention Mechanism} \label{sec:method-mam}

The second part of the branches consists of a modified neural attention \textcolor{black}{mechanism} on top of the output hidden states produced by the FCGRU layer. Attention strategies~\cite{Bahdanau14,Luong15,BritzGL17} are widely used in automatic signal processing (1D signal or language) as they allow to join together the information extracted by the RNN model at different time stamps via a convex combination of the input sources. Attention was initially designed in the context of Neural Machine Translation using sequence to sequence (seq2seq) models~\cite{SutskeverVL14}, to deal with long sequences sentences which decrease such models performances. The traditional seq2seq models are composed of a RNN encoder and decoder. A context (thought) vector summarizing (encoding) the encoder states is then used by the decoder to predict the output sequences, discarding the encoder intermediate states. In attention models, instead of throwing away those intermediate states, \textcolor{black}{the model uses them} to construct the context vector required by the decoder. The hypothesis is that all the states are not necessary relevant when trying to predict a specific target sequence (word). Therefore, a score (attention weight) is computed for each intermediate state as the decoder will "pay more attention" to states which get high weights in the target sequence predicting. The attention weights are \textcolor{black}{commonly} computed using a $SoftMax$ function so that their values range is [0,1] and their sum is equal to 1 providing at the same time a probabilistic interpretation. Finally, the context vector is computed by weighted each intermediate state with the attributed weight.

However, in the remote sensing time series classification context, where no decoder is required, using a $SoftMax$ function forcing the sum of weights to 1 may not be fully beneficial for the attention model. In fact, considering a specific time-series classification task where almost all the time stamps are relevant for the problem, \textcolor{black}{the use of a SoftMax function to compute attention weights} will squash towards zero the attention weights since their sum should be one and finally the attention combination may not be efficient as expected. \textcolor{black}{Relaxing} this constraint could thus help to better weight the relevant time stamps \textcolor{black}{independently}. Therefore, in our attention formulation, we attempt to address this point by substituting the $SoftMax$ function with a Hyperbolic Tangent function in the attention weight computation. The motivation behind the $tanh$ neural attention apart from \textcolor{black}{relaxing} the sum constraint is that the learned weights will be in a wider range i.e. [-1,1], also allowing negative scores to be mapped as negative weights. The equations below describe the $tanh$ attention formulation \textcolor{black}{we introduce}:
\begin{equation}
score = tanh(H \cdot W + b) \cdot u \label{eqn:att1}
\end{equation}
\begin{equation}
\lambda = tanh(score) \label{eqn:att2}
\end{equation}
\begin{equation}
feat = \sum_{i=1}^{N} \lambda_i h_{t_{i}} \label{eqn:att3}
\end{equation}

\noindent where $H \in \mathbb{R}^{N,d}$ is a matrix obtained by vertically stacking all hidden state vectors $h_{t_i} \in \mathbb{R}^{d}$ learned at $N$ different timestamps by the FCGRU; $\lambda \in \mathbb{R}^{d}$ is the attention weight vector traditionally computed by a $SoftMax$ function which we replaced by a $tanh$ function; matrix $W \in \mathbb{R}^{d,d}$ and vectors $b, u \in \mathbb{R}^{d}$ are parameters learned during the process.

\textcolor{black}{The described attention mechanism is employed over the FCGRU outputs (hidden states) in the radar and optical branches to generate per-source features ($feat_{rad}$ and $feat_{opt}$). Such features encode the temporal information related to the input source. Furthermore, the per-source hidden states are concatenated and an additional attention mechanism is employed over them to generate fused features ($feat_{fused}$). Such fused features encode both temporal information and complementarity of radar and optical sources. Thus, the architecture involves learning three sets of attention weights:} $\lambda_{rad}$, $\lambda_{opt}$ and $\lambda_{fused}$ which refers respectively to the attention mechanisms employed over the radar hidden states, the optical hidden states and the concatenated ones. 

\subsection{Feature combination}

Once each set of features has been yielded, they are directly leveraged to perform the final land cover classification. The combination process involves \textcolor{black}{three} classifiers: one classifier on top of the fused features ($feat_{fused}$) and two auxiliary classifiers, \textcolor{black}{one for each source} ($feat_{rad}$ and $feat_{opt}$). Auxiliary classifiers~\cite{HouLW17,INTERDONATO201991,IENCO201911} are used to strengthen the complementarity as well as the discriminative power of the per-source learned features. The goal of this extra classifiers is to stress the fact that the learned features need to be discriminative alone i.e. independently from each other. 

Then, the cost function associated to the optimization of the three classifiers is: 
\begin{equation}
\footnotesize
L_{total} = 0.5 * L(feat_{rad}) + 0.5 * L(feat_{opt}) + L(feat{fused}) \label{eqn:cost}
\end{equation}
\noindent where $L(feat)$ is the loss function (in our case the categorical Cross-Entropy) associated to the classifier fed with the features $feat$. 

The contribution of each auxiliary classifier was empirically weighted by 0.5 to enforce the discriminative power of the per-source learned features while privileging the fused features in the combination. The final land cover class is derived combining the three classifiers with the same weight schema employed in the loss function computation:
\begin{equation}
score =  0.5  \times score_{rad} + 0.5 \times score_{opt} + score_{fused}
\end{equation}
where $score_{fused}$, $score_{rad}$ and $score_{opt}$ are the predictions of the fused classifier, the classifier considering the radar SITS and that considering the optical SITS, respectively.

\subsection{Hierarchical pretraining strategy}
Considering real-world scenario, when field campaigns are performed, information on a plot can be collected considering multiple level of details. For instance, given a plot, the expert collects the ground truth, filling firstly the vegetation land cover, then the crop type and finally the crop. 

\begin{figure}[!htbp]
\centering
\includegraphics[width=.6\columnwidth]{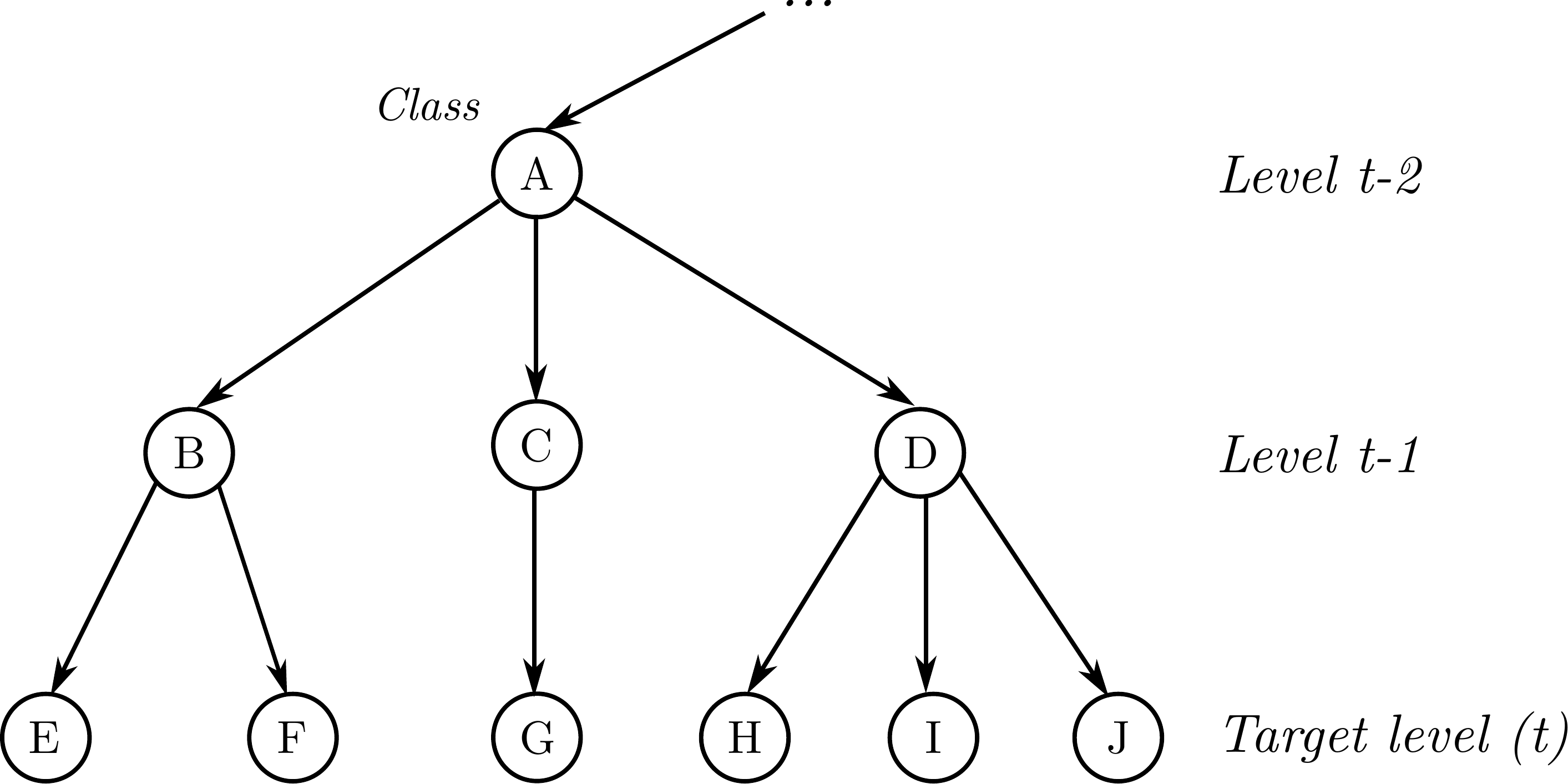}
\caption{ \label{fig:hier} Representation of a land cover class hierarchical organization}
\end{figure}

Therefore, considering the simplest to the most complex level, we can derive a hierarchical organization about land cover classes, as illustrated in Figure~\ref{fig:hier}, which could be fruitful to consider in the land cover classification process. 

With the aim to leverage such a precious knowledge, the~\method{} architecture was trained in a hierarchical manner~\textcolor{black}{exploiting such taxonomic organization of the land cover classes.
The training is repeated for each level of the hierarchy, from the more general one (the most simple) to the most specific one (the target classification level). }

Specifically, we start the model training on the~\textcolor{black}{ highest level of the hierarchy} and subsequently, we continue the training on the next level reusing the previous learned weights for the whole architecture, excepting \textcolor{black}{the weights associated to the output layer since level-specific. This process is performed until we reach the target level.} New classifiers \textcolor{black}{(output layers) were} trained for each level \textcolor{black}{of the hierarchy. To sum up, such hierarchical pretraining strategy allows the model to focus firstly on easy classification problems (i.e. Crops vs Non Crops) and, step by step, the network behavior is adapted smoothly to deal with classification problems with increasing complexity levels. In addition, this process allows the classification model to tackle the classification at the target level (the most specific level of the land cover hierarchy) integrating a kind of prior knowledge on the task instead of addressing it completely from scratch.}

\section{Experimental evaluation} \label{sec:exp}

In this section, we present and discuss the experimental results obtained on the study sites introduced in Section~\ref{sec:data}. We carried out several experimental analysis in order to provide a deep assessment of the \method{} behaviour:
\begin{itemize}
    \item an in-depth evaluation of the quantitative performances of the \method{} with respect to several other competitors; 
    \item \textcolor{black}{an ablation study on sources (Sentinel-1/Sentinel-2) as well as on the different components of the \method{} architecture to characterize the interplay among them;}
    \item a qualitative analysis of land cover maps considering the \method{} and its competitors;
    \item an inspection of the attention parameters learnt by our architecture, in order to investigate to what extent such side information can contribute to the model interpretability.
\end{itemize}

\subsection{Experimental settings}
To assess the quality of \method{}, we \textcolor{black}{chosen} as competitors common machine learning techniques \textcolor{black}{which are the de facto baselines in the field of remote sensing, commonly employed to deal with SITS data~\cite{Erinjery2018345}:} Random Forest (RF) and Support Vector Machine (SVM). \textcolor{black}{In addition, we also consider} a Multi Layer Perceptron neural net (MLP). The competing methods were run over the concatenation of the \textcolor{black}{different information sources:} Sentinel-1 and Sentinel-2 SITS. 
\textcolor{black}{We optimize the model parameters via train/validation procedure~\cite{IencoGDM17}.}
The settings are reported in Table~\ref{tab:settings}.
The datasets were split into training, validation and test set with an object proportion of 50\%, 20\% and 30\% respectively. The values were normalized per band (resp. indices) considering the time series, in the interval $[0,1]$. Training data are used to learn the model while validation data are exploited for model selection. 
Finally, the model achieving the best accuracy on the validation set was employed to classify the test set. We imposed that segments belonging to the same ground truth polygon before the spatial intersection (see Section~\ref{subsec:GT}) were assigned exclusively to one of the \textcolor{black}{data partitions} (training, validation or test) to avoid a possible spatial bias in the evaluation procedure. The assessment of the classification performances was done considering \textit{Accuracy} (global precision), \textit{F1 Score} and \textit{Kappa} metrics. Since the model performances may vary depending on the split of the data due to simpler or more complex samples involved in the different \textcolor{black}{partition}, all metrics were averaged over ten random splits following the strategy previously reported. Experiments were carried out on a workstation with an Intel (R) Xeon (R) CPU E5-2667 v4@3.20Ghz with 256~GB of RAM and four TITAN X GPU. The neural architectures were implemented using the Python Tensorflow library, \textcolor{black}{while machine learning algorithms implementation were obtained from the} Python Scikit-learn library~\cite{sklearn}. 

\begin{table}[!htbp]
\caption{Hyper-parameters and corresponding value or ranges of all the competing methods}
\label{tab:settings}
\centering
 \begin{tabular}{|c|c|c|}
  \hline
  Method & Hyper-Parameter & Value or Range \\
  \hline
  \multirow{4}*{\method} & Epoch number & 2000 per level \\
  & FCGRU units & 512 \\ 
  & FC1 units &  64 \\ 
  & FC2 units &  128 \\ 
  \hline
  \multirow{3}*{MLP} & Epoch number & 2000 \\ 
  & Hidden units & 512 \\ 
  & Hidden layers & 2 \\
  \hline
  \multirow{2}*{\method} & Batch size & 32 \\
  & Dropout rate & 0.4 \\
  \multirow{2}*{and MLP} & Optimizer & Adam~\cite{KingmaB14} \\ 
  & Learning rate & \num{1e-4} \\ 
  \hline
  \multirow{3}*{RF} & Tree number & \{100, 200, 300,400,500\} \\ 
  & Maximum depth & \{20,40,60,80,100\} \\ 
  & Maximum features & \{'sqrt', 'log2', None\} \\ 
  \hline
  \multirow{3}*{SVM} & Kernel & \{'linear', 'poly', 'rbf', 'sigmoid'\} \\ 
  & Gamma & \{0.25,0.5,1,2\} \\ 
  & Penalty & \{0.1, 1, 10\} \\ 
  \hline
 \end{tabular}
\end{table}

\subsection{Comparative analysis}
In this part, we compare the results obtained by the different competing methods, considering their overall and per-class scores. 

\subsubsection{General behavior}
Table~\ref{tab:avg} reports the averaged results obtained for RF, SVM, MLP and \method{} on the \textit{Reunion island} and the \textit{Senegalese groundnut basin} site. Considering the averaged performances, we can state that the proposed method outperforms its competitors on both study sites. However, the performance gaps are better pronounced on the \textit{Reunion island} dataset than the \textit{Senegalese} site one. This behavior may be due to the fact that the \textit{Reunion island} dataset represents a more rich benchmark (about 8 times) than the \textit{Senegalese} dataset in terms of ground truth samples. In fact, it is known that deep learning models get more and more accurate when learning from \textcolor{black}{huge volumes} of data. 
Concerning the other competing methods 
i.e. RF and SVM, they \textcolor{black}{achieve similar} scores on the \textit{Reunion island} while SVM surpasses RF on the \textit{Senegalese} site. On this latter benchmark, the SVM algorithm \textcolor{black}{demonstrates to be well suited for dataset characterized by a limited set of labeled samples}. As regards the MLP competitor, 
it achieves lower scores than \method{} on the \textit{Reunion island} while scores are comparable on the \textit{Senegalese} site. 
It should be noted that the relatively better performance obtained on the \textit{Senegalese} site compared to the \textit{Reunion island} (\textcolor{black}{90.78 vs. 79.66})
may come from the topography of the study sites. In fact, \textit{Reunion island} is characterized by a rugged topography while the \textit{Senegalese} site is essentially flat. Relief effects like shadow or orientation can induce biases in the discrimination of land cover classes impacting much more the \textit{Reunion island}~\cite{IENCO201911}. 


\begin{table}[!htbp]
\caption{F1 score, Kappa and Accuracy considering the different methods (RF, SVM, MLP and \method{}) on each study site (results averaged over ten random splits)}
\label{tab:avg}
\centering
 \begin{tabular}{|c|c|c|c|c|}
  \hline
  Site & Method & F1 score & Kappa & Accuracy \\   \hline
  \multirow{4}*{Reunion} & RF & 75.62 $\pm$ 1.00 & 0.726 $\pm$ 0.011 & 75.75 $\pm$ 0.98 \\
  & SVM & 75.34 $\pm$ 0.88 & 0.722 $\pm$ 0.010 & 75.39 $\pm$ 0.89 \\
  & MLP & 77.96 $\pm$ 0.70 & 0.752 $\pm$ 0.008 & 78.03 $\pm$ 0.66 \\
  & \method & \textbf{79.66} $\pm$ \textbf{0.85} & \textbf{0.772} $\pm$ \textbf{0.009} & \textbf{79.78} $\pm$ \textbf{0.82} \\
  \hline 
  \multirow{4}*{Senegal} & RF & 86.31 $\pm$ 0.91 & 0.828 $\pm$ 0.012 & 86.35 $\pm$ 0.90 \\
  & SVM & 89.95 $\pm$ 0.85 & 0.875 $\pm$ 0.011 & 89.96 $\pm$ 0.85 \\
  & MLP & 90.05 $\pm$ 0.56 & 0.876 $\pm$ 0.007 & 90.07 $\pm$ 0.57 \\
  & \method & \textbf{90.78} $\pm$ 1.03 & \textbf{0.885} $\pm$ 0.013 & \textbf{90.78} $\pm$ 1.03 \\
  \hline
 \end{tabular}
\end{table}

\subsubsection{Per-class analysis}

Tables~\ref{tab:per-reunion} and~\ref{tab:per-senegal} report the per-class F1 scores obtained by the different methods on the \textit{Reunion island} and the \textit{Senegalese} site, respectively. 
Concerning the \textit{Reunion} site, we can observe that~\method{} achieves the best performances on the majority of land cover classes excepted some classes where other competing methods i.e. RF or MLP obtained slightly better scores \textcolor{black}{that are still comparable to the ones achieved by our framework.}  It is worth noting how the proposed method outperforms its competitors particularly on agricultural/vegetation classes i.e. classes $0$--\textit{Sugar cane}, $1$--\textit{Pasture and fodder}, $2$--\textit{Market gardening}, $3$--\textit{blackhouse crops} and $4$--\textit{Orchards}. The best gain (5 points from the best competitor) is obtained on $1$--\textit{Pasture and fodder} class. This particular efficiency on such classes suggest that the \method{} architecture is well suited to deal with \textcolor{black}{the} temporal dependencies characterizing these land cover classes. 
As regards the \textit{Senegalese} site, the per-class \method{} performances are more moderate. It achieves the best scores on 4 land cover classes over 9 i.e. $1$--\textit{Fallows}, $2$--\textit{Ponds}, $7$--\textit{Cereals} and $8$--\textit{Legumes} while other competing methods outperformed its results especially on the $6$--\textit{Valley} class. Nonetheless, it should be remarked that also in this case, \method{} obtained the best results on \textcolor{black}{land cover classes which exhibit a time-varying behavior.} It is common to observe natural vegetation activity on fallows areas; ponds appear during the rainy season while cereals and legumes follow crop growth cycle. These findings are inline with the previous observations made on the \textit{Reunion island} and confirm the \textcolor{black}{fact that the} proposed method \textcolor{black}{is capable to fruitfully} leverage temporal dependencies \textcolor{black}{to made its decision.}

\begin{table}[!htbp]
\centering
\footnotesize
\caption{Per-Class \textit{F1 score} for the \textit{Reunion island} (average over ten random splits) \label{tab:per-reunion}}
\begin{tabular}{|c|c|c|c|c|c|c|c|c|c|c|c|c|c|c|c|}
	\hline
 \textbf{Method} & \rotatebox{90}{$0$--\textit{Sugar cane}} & \rotatebox{90}{$1$--\textit{Pasture and fodder}} & \rotatebox{90}{$2$--\textit{Market gardening}} & \rotatebox{90}{$3$--\textit{blackhouse crops}} & \rotatebox{90}{$4$--\textit{Orchards}} & \rotatebox{90}{$5$--\textit{Wooded areas}} & \rotatebox{90}{$6$--\textit{Moor and Savannah}} & \rotatebox{90}{$7$--\textit{Rocks and bare soil}} & \rotatebox{90}{$8$--\textit{Relief shadows}} & \rotatebox{90}{$9$--\textit{Water}} & 
 \rotatebox{90}{$10$--\textit{Urbanized areas}} \\  \hline
RF & 83.34 & 76.13 & 66.39 & 47.85 & 67.16 & \textbf{83.10} & 75.09 & 81.81 & \textbf{93.51} & 78.44 & 74.54 \\ 
SVM & 83.61 & 74.46 & 64.12 & 50.69 & 68.40 & 82.41 & 75.28 & 79.76 & 87.44 & 75.88 & 77.31 \\
MLP & 85.95 & 76.92 & 69.25 & 52.87 & 70.72 & 82.36 & \textbf{78.16} & 82.57 & 90.00 & 84.91 & \textbf{79.24} \\
\method & \textbf{88.67} & \textbf{81.26} & \textbf{71.73} & \textbf{53.43} & \textbf{72.19} & 82.62 & 77.44 & \textbf{85.71} & 90.61 & \textbf{88.77} & 79.05 \\
\hline
\end{tabular}
\end{table}

\begin{table}[!htbp]
\centering
\footnotesize
\caption{Per-Class \textit{F1 score} for the \textit{Senegalese} site (average over ten random splits) \label{tab:per-senegal}}
\begin{tabular}{|c|c|c|c|c|c|c|c|c|c|c|c|c|c|}
	\hline
 \textbf{Method} & \rotatebox{90}{$0$--\textit{Bushes}} & \rotatebox{90}{$1$--\textit{Fallows}} & \rotatebox{90}{$2$--\textit{Ponds}} & \rotatebox{90}{$3$--\textit{Bare soils}} & \rotatebox{90}{$4$--\textit{Villages}} & \rotatebox{90}{$5$--\textit{Wet areas}} & \rotatebox{90}{$6$--\textit{Valley}} & \rotatebox{90}{$7$--\textit{Cereals}} & \rotatebox{90}{$8$--\textit{Legumes}} \\  \hline
RF & 73.10 & 81.03 & 78.34 & 92.54 & 97.86 & 83.67 & \textbf{92.29} & 81.49 & 82.99 \\ 
SVM & 82.08 & 82.95 & 87.08 & 93.84 & \textbf{99.04} & \textbf{91.12} & 92.04 & 86.64 & 87.09 \\ 
MLP & \textbf{82.37} & 84.04 & 84.96 & \textbf{93.90} & 98.98 & 90.51 & 90.27 & 86.97 & 87.15 \\
\method & 80.68 & \textbf{86.20} & \textbf{89.12} & 91.84 & 98.78 & 89.84 & 88.52 & \textbf{87.71} & \textbf{89.34} \\ 
\hline
\end{tabular}
\end{table}

\begin{figure}[!htbp]
\centering
\subfloat[RF\label{fig:cm-reunion1}]{\includegraphics[width=0.5\linewidth]{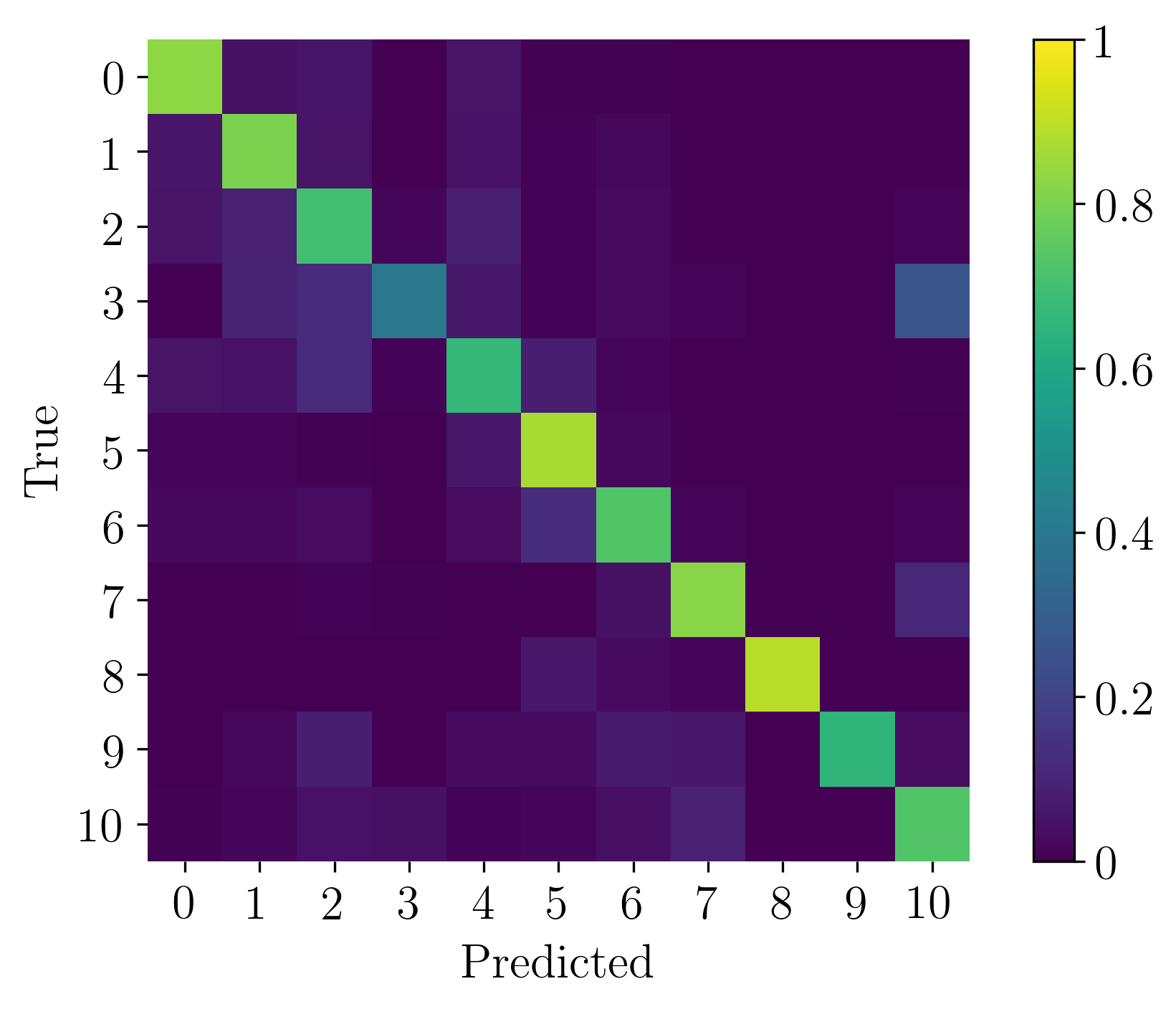}}
\hfill
\subfloat[SVM\label{fig:cm-reunion2}]{\includegraphics[width=0.5\linewidth]{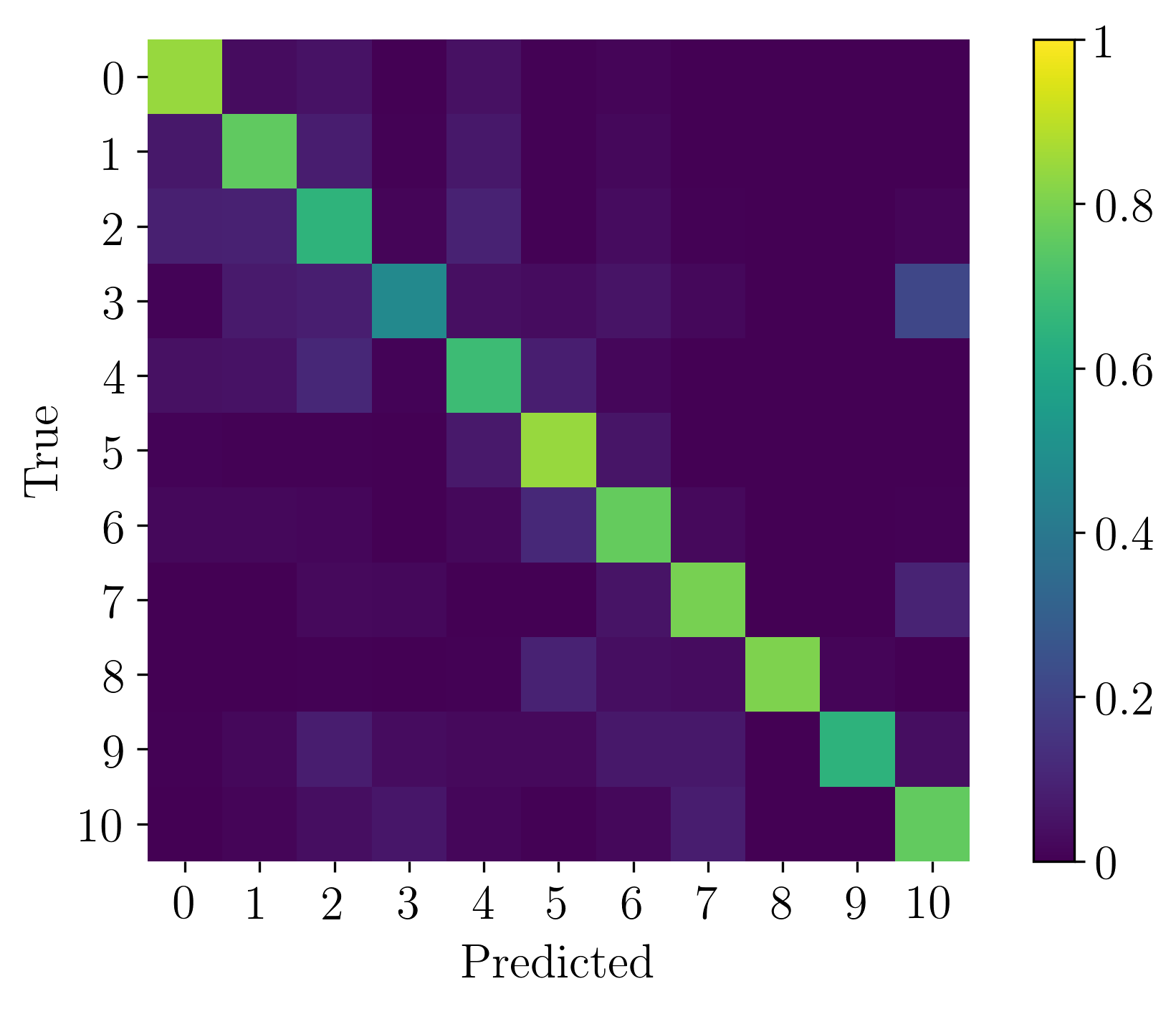}}

\subfloat[MLP\label{fig:cm-reunion3}]{\includegraphics[width=0.5\linewidth]{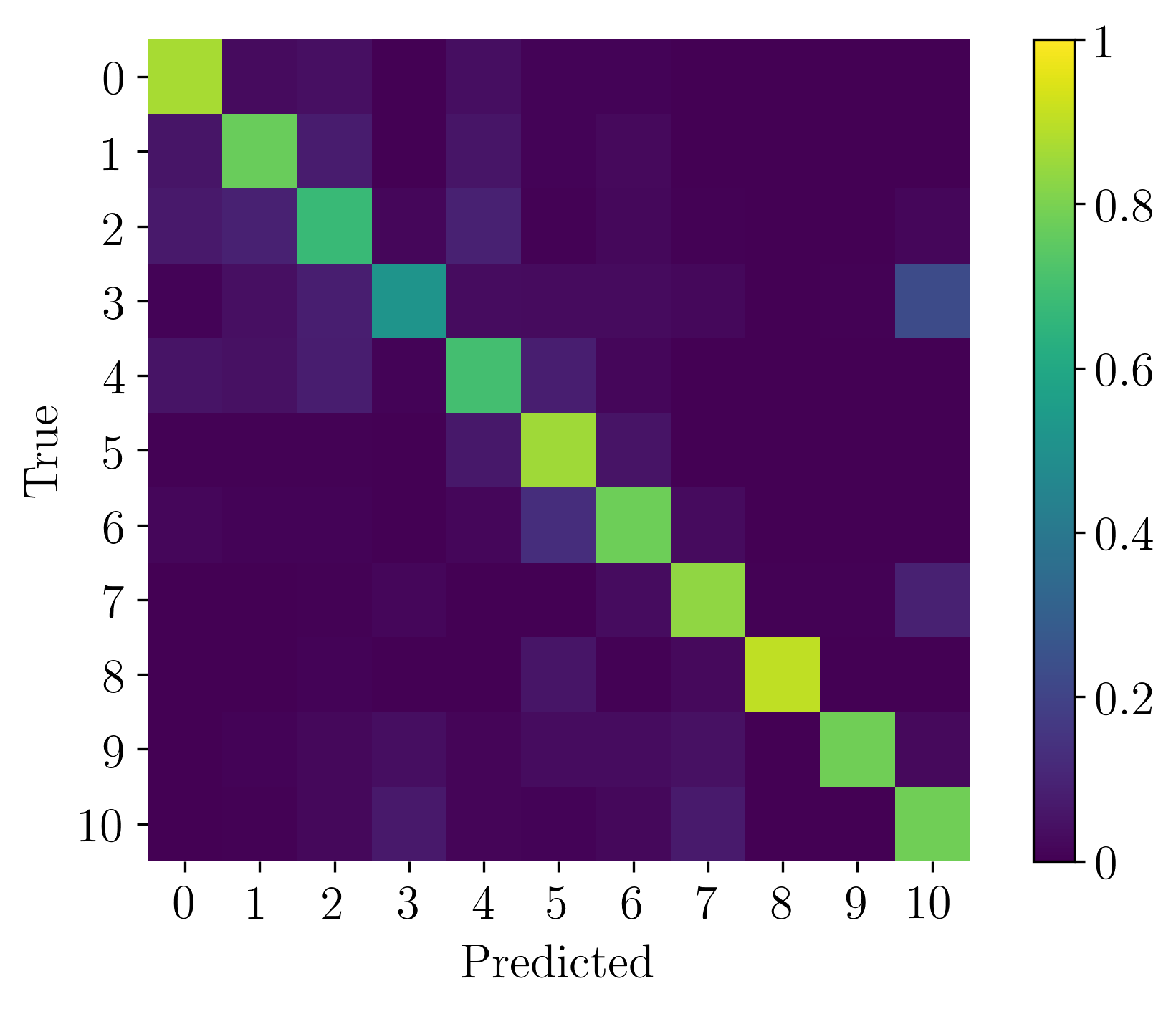}}
\hfill
 \subfloat[\method{}\label{fig:cm-reunion4}]{\includegraphics[width=0.5\linewidth]{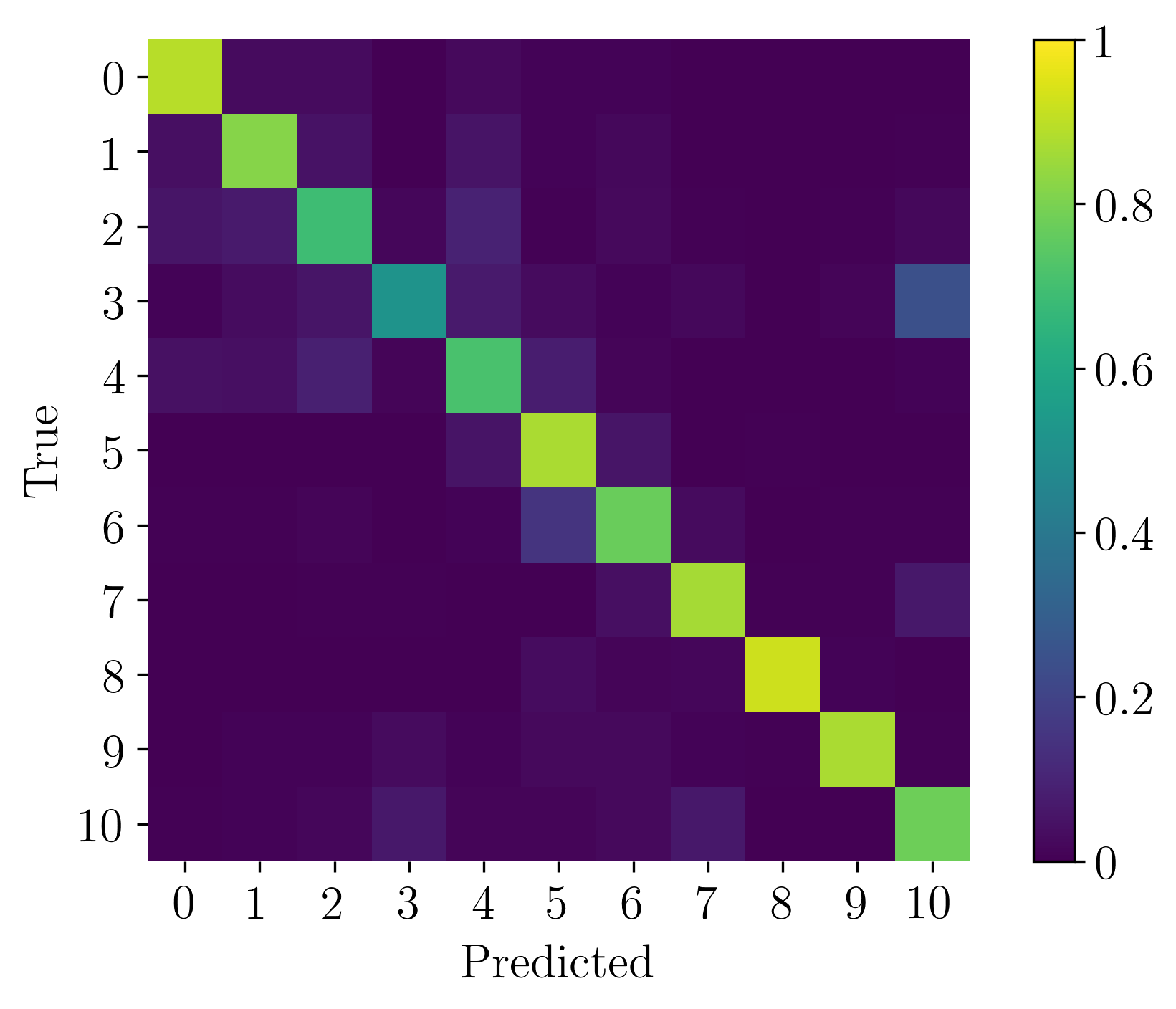}}
\caption{Confusion Matrices of the land cover classification produced by (a) RF, (b) SVM, (c) MLP and (d) \method{} on the \textit{Reunion island}.}
\label{fig:cm-reunion}
\end{figure}

\begin{figure}[!htbp]
\centering
\subfloat[RF\label{fig:cm-senegal1}]{\includegraphics[width=0.5\linewidth]{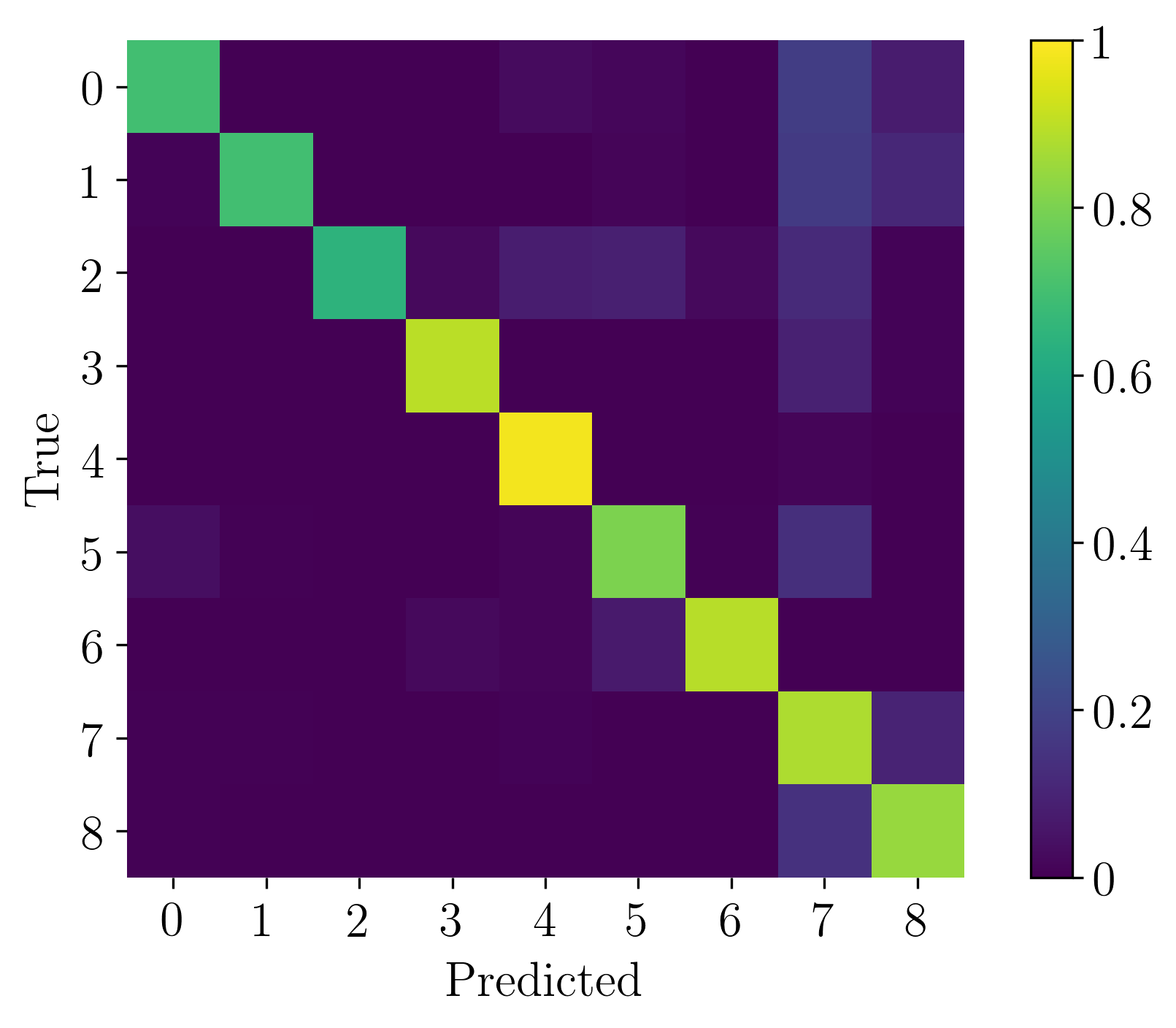}}
\hfill
\subfloat[SVM\label{fig:cm-senegal2}]{\includegraphics[width=0.5\linewidth]{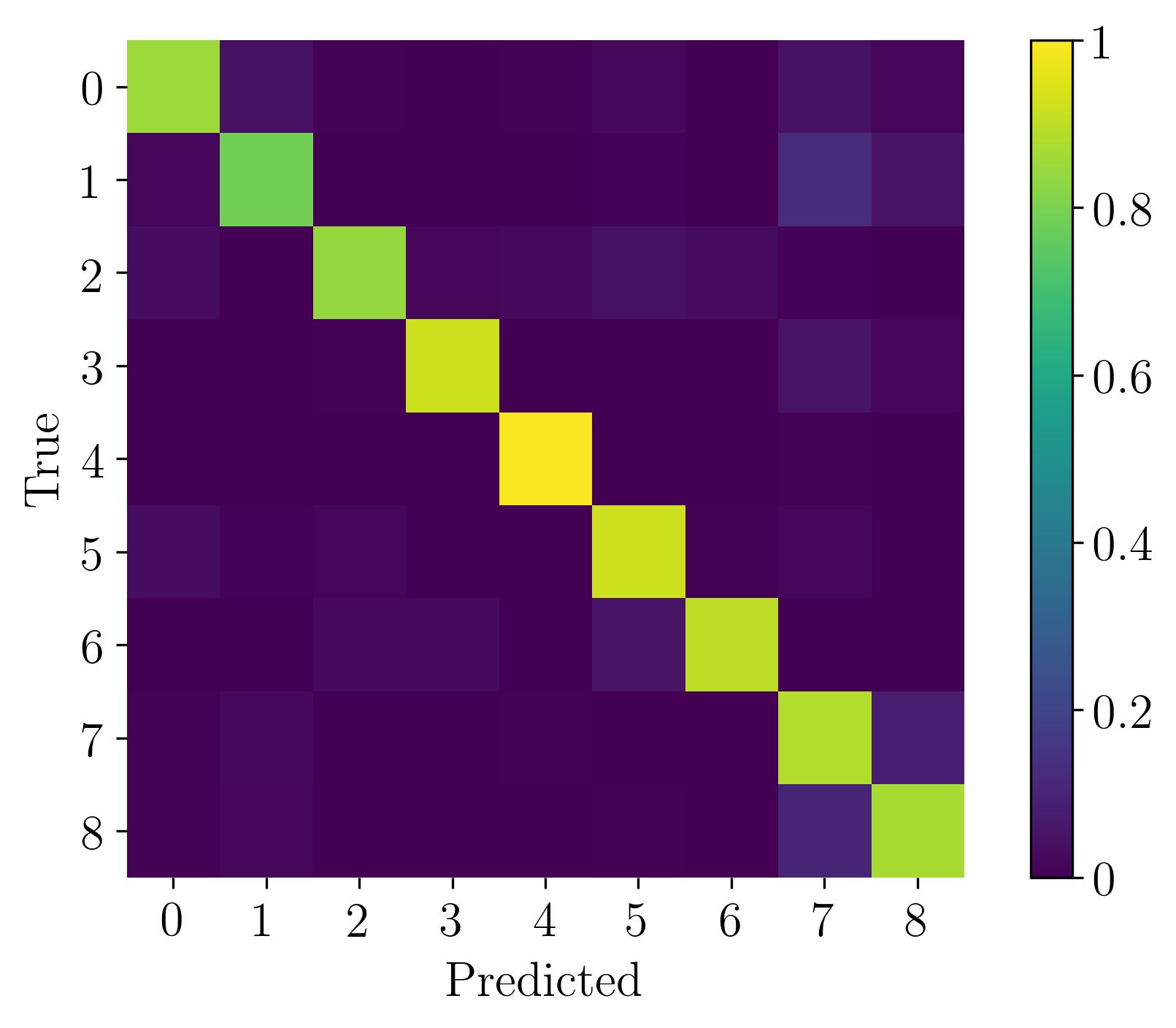}}
\hfill
\subfloat[MLP\label{fig:cm-senegal3}]{\includegraphics[width=0.5\linewidth]{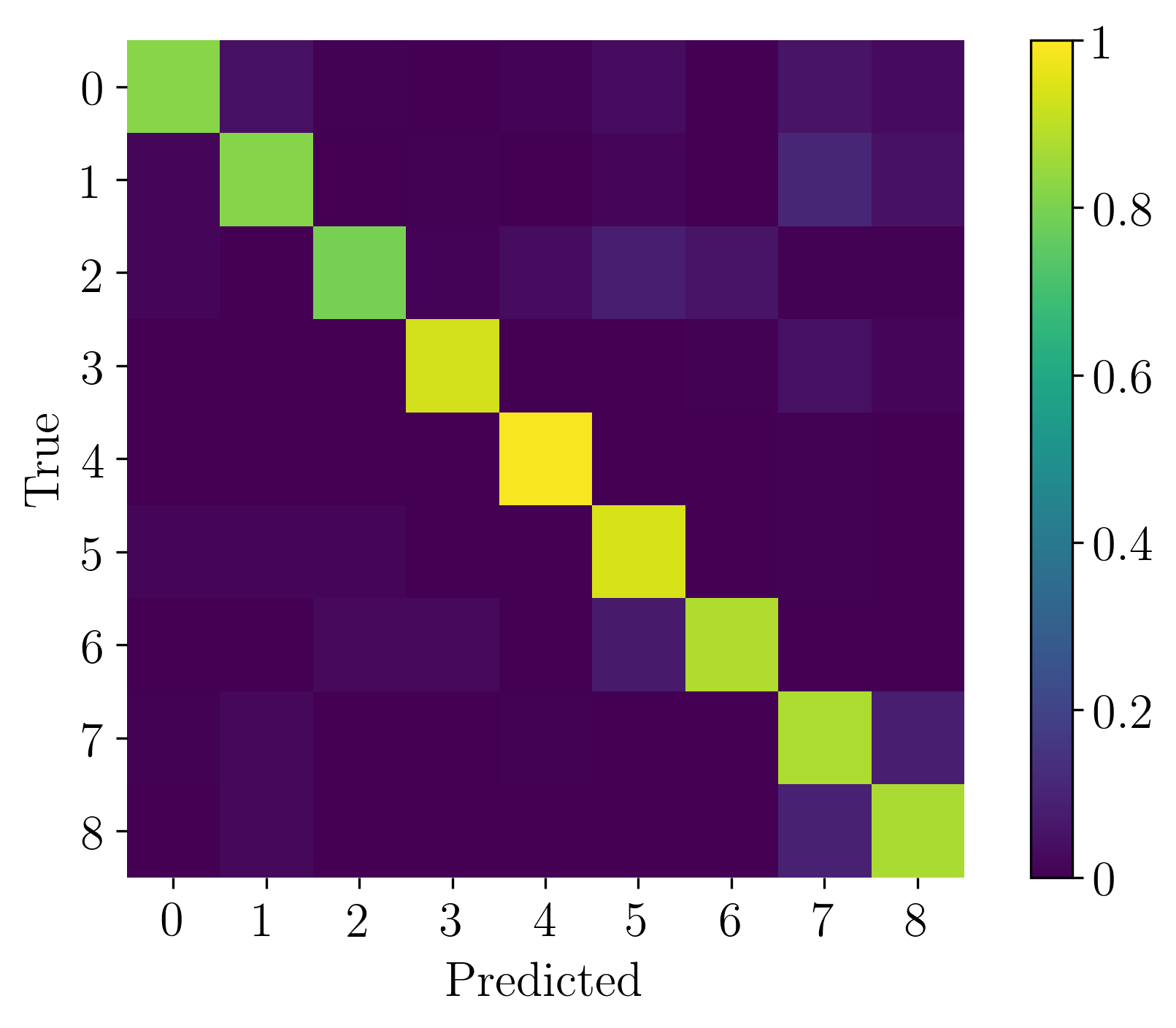}}
\hfill
 \subfloat[\method{}\label{fig:cm-senegal4}]{\includegraphics[width=0.5\linewidth]{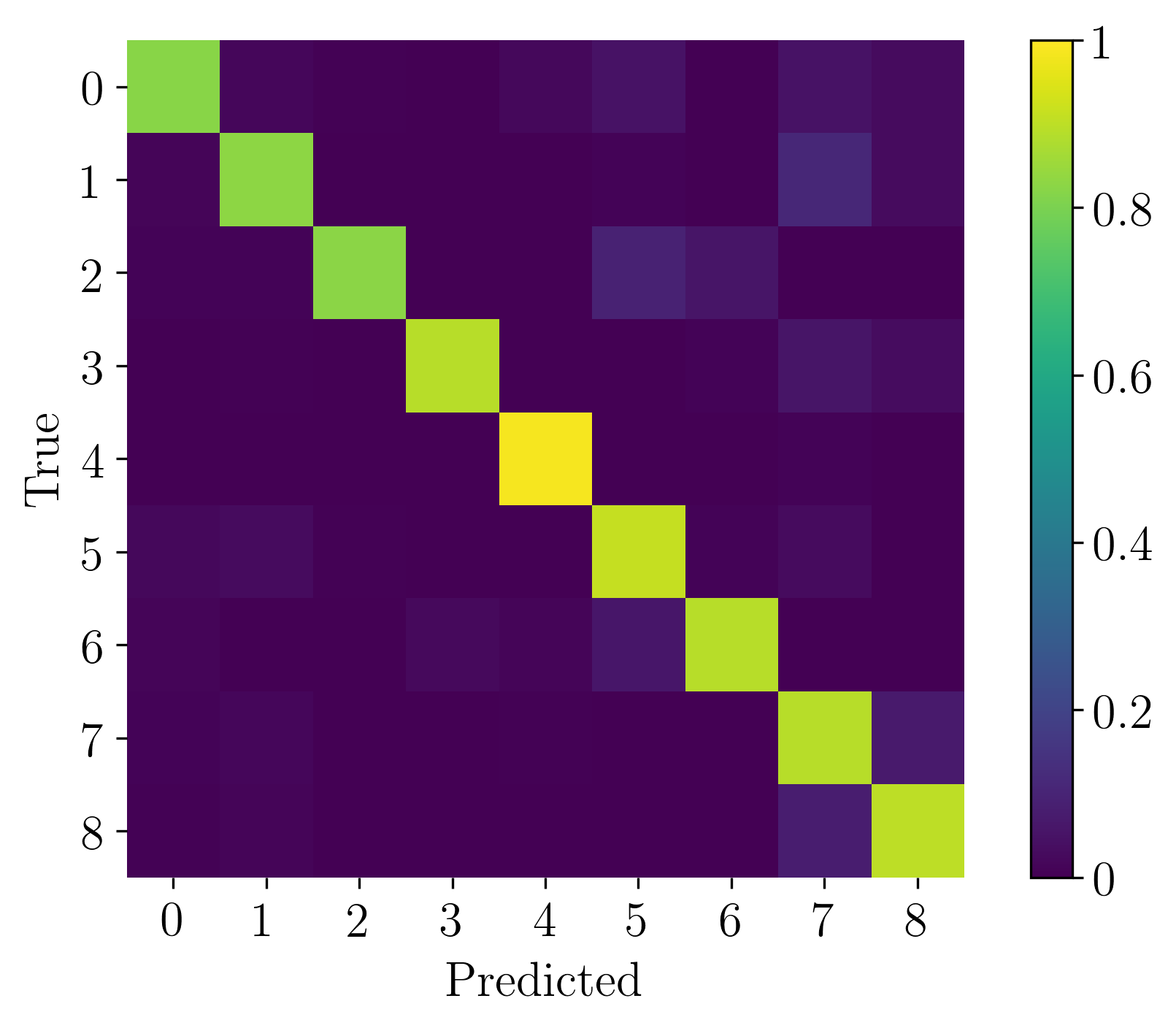}}
\caption{Confusion Matrices of the land cover classification produced by (a) RF, (b) SVM, (c) MLP and (d) \method{} on the \textit{Senegalese} site.}
\label{fig:cm-senegal}
\end{figure}

To go further with the per-class analysis, we also investigate the confusions matrices of each method on the \textcolor{black}{two} study sites. Concerning the \textit{Reunion island} (Figure~\ref{fig:cm-reunion}), \textcolor{black}{all the methods exhibit similar behaviors. This is particularly evident}
between $3$--\textit{blackhouse crops} and $10$--\textit{Urbanized areas} class even if confusion are reduced from RF (Figure \ref{fig:cm-reunion1}) to \method{} as can be observed (Figure~\ref{fig:cm-reunion4}). 
\textcolor{black}{Overall, the per-class analysis is coherent with the findings we got in the previous analysis.}
Apropos of the \textit{Senegalese} site (Figure \ref{fig:cm-senegal}), confusions vary sensibly regarding the different methods. RF (Figure \ref{fig:cm-senegal1}) exhibits more confusions on $0$--\textit{Bushes} and $1$--\textit{Fallows} classes which are highly misclassified with  $7$--\textit{Cereals} class and a little bit less with $8$--\textit{Legumes} class while $2$--\textit{Ponds} are often confused with $5$--\textit{Wet areas}. The other competitors tend to reduce these confusions as also evidenced \textcolor{black}{by the confusion matrix colours.} 

\subsection{Ablation analysis}
Here, we conduct several stages of ablation experiments considering the multi-source SITS and the architecture components. In addition, we provide an assessment of the NDVI index helpfulness, used as extra optical descriptor, for the \method{} architecture.

\subsubsection{Ablation on multi-source SITS}

In this stage of experiments, we consider only one source of time series (radar or optical) to perform the land cover classification. We name \method(S1) (resp. \method(S2) ) the ablation of our model considering the radar (resp. optical) branch. To better figure out how each source of SITS is leveraged, we also report the competing method performances (RF, SVM and MLP) in this per-source ablation analysis. Their variants are named in the same manner.

Regarding the results reported in Table \ref{tab:source}, \textcolor{black}{the radar time series has a specific behavior for each of the considered study sites.} If radar signal is quite discriminating in the \textit{Senegalese} site, this is not really the case for the \textit{Reunion island} \textcolor{black}{considering how poorly the competing methods trained on the radar SITS perform, especially the SVM algorithm.} 
Thus on the \textit{Reunion island}, non temporal based models i.e. (RF, SVM and MLP) perform slightly worst or equally on the concatenation of 2 sources than learning only with optical data. However, \method{} performs better when combining the \textcolor{black}{two time series} sources. This behavior suggests that \method{} is able to better leverage the complementarity between radar and optical data than its competitors. This behavior is also evident in the \textit{Senegalese} site where all competing methods perform better with both sources but \method{} is the one that performs the best. For the rest, considering the \textcolor{black}{two} study sites, there is no trend on which competing methods better deal with radar or optical time series than others. However, we have observed on both sites that SVM algorithm \textcolor{black}{seems not well suited to exploit radar information.} 

\begin{table}[!htbp]
\caption{F1 score, Kappa and Accuracy of the different methods (RF, SVM, MLP and \method{}) considering the per-source ablation analysis on the study sites (results averaged over ten random splits). Refer to Table \ref{tab:avg} for the results of SITS combination.}
\label{tab:source}
\centering
 \begin{tabular}{|c|c|c|c|c|}
  \hline
  Site & Classifier & F1 score & Kappa & Accuracy \\   \hline
  \multirow{8}*{Reunion} & RF(S1) & \textbf{36.77} $\pm$ 0.93 & \textbf{0.291} $\pm$ 0.011 & \textbf{37.85} $\pm$ 0.95 \\
  
  & SVM(S1) & 6.56 $\pm$ 0.36 & 0.018 $\pm$ 0.009 & 16.85 $\pm$ 0.53 \\
  & MLP(S1) & 34.93 $\pm$ 1.42 & 0.271 $\pm$ 0.016 & 36.01 $\pm$ 1.39 \\
  & \method(S1) & 31.80 $\pm$ 1.10 & 0.231 $\pm$ 0.011 & 32.39 $\pm$ 1.04 \\
  & RF(S2) & 76.24 $\pm$ 0.59 & 0.732 $\pm$ 0.007 & 76.32 $\pm$ 0.63 \\
  & SVM(S2) & 75.55 $\pm$ 0.80 & 0.724 $\pm$ 0.009 & 75.60 $\pm$ 0.80 \\
  & MLP(S2) & 77.95 $\pm$ 0.69 & 0.751 $\pm$ 0.008 & 77.98 $\pm$ 0.73 \\
  & \method(S2) & \textbf{78.69} $\pm$ 0.95 & \textbf{0.761} $\pm$ 0.010 & \textbf{78.79} $\pm$ 0.91 \\
  \hline 
  \multirow{8}*{Senegal} & RF(S1) & 75.71 $\pm$ 1.03 & 0.703 $\pm$ 0.013 & 76.56 $\pm$ 1.00 \\
  & SVM(S1) & 71.27 $\pm$ 0.82 & 0.653 $\pm$ 0.010 & 72.82 $\pm$ 0.78 \\
  & MLP(S1) &  \textbf{78.96} $\pm$ 1.28 & \textbf{0.738} $\pm$ 0.015 & \textbf{79.05} $\pm$ 1.23 \\
  & \method(S1) & 77.42 $\pm$ 1.33 & 0.721 $\pm$ 0.016 & 77.63 $\pm$ 1.27 \\ 
  & RF(S2) & 84.51 $\pm$ 1.17 & 0.806 $\pm$ 0.015 & 84.60 $\pm$ 1.17 \\ 
  & SVM(S2) & \textbf{88.64} $\pm$ 0.47 & \textbf{0.858} $\pm$ 0.006 & \textbf{88.63} $\pm$ 0.45 \\ 
  & MLP(S2) & 88.38 $\pm$ 0.61 & 0.855 $\pm$ 0.008 & 88.40 $\pm$ 0.62 \\
  & \method(S2) & 87.56 $\pm$ 1.33 & 0.845 $\pm$ 0.017 & 87.55 $\pm$ 1.33 \\ 
  \hline
 \end{tabular}
\end{table}

\subsubsection{Ablation on architecture components}

In this part, we investigate the interplay among the different components of \method{} and 
we disentangle their benefits in the architecture. We consider both time series (radar and optical) but excluding one of the following components at a time: the \textcolor{black}{three} attention mechanisms involved in the architecture (naming $NoAtt$), the hierarchical pretraining process (naming $NoHierPre$) and the enrichment step in the FCGRU cell which is equivalent to use a GRU cell (naming $NoEnrich$). We also \textcolor{black}{investigate} the use of the traditional $SoftMax$ attention mechanism instead of the modified one. This variant is named $SoftMaxAtt$. Results are reported in Table \ref{tab:component}.

Concerning the use of attention mechanisms or not ($NoAtt$, $SoftMaxAtt$ and \method), we can observe how these components contribute to the final classification performances on both study sites, more on the \textit{Reunion island} (about 2 points of improvement) than the \textit{Senegalese} site (approximately 1 point). We can also note that $SoftMax$ attention performs relatively similar than the non use of attention mechanisms and lower than the $tanh$ attentions confirming our hypothesis that \textcolor{black}{relaxing the constraint that the attention weights may sum to 1} in the attention process is more suitable for remote sensing context. As regards the use of the hierarchical pretraining process ($noHierPre$ and \method), we can note here the added value of such step on both study sites obtaining more than 1 point of improvement. These results \textcolor{black}{seem to underline that involving domain specific knowledge in the pretraining process of neural networks can} improve the final classification performances. Finally, the new FCGRU cell compared to the GRU cell ($noEnrich$) performs better in both study sites, however it seems to be more efficient in the \textit{Senegalese} site. 

\begin{table}[!htbp]
\caption{F1 score, Kappa and Accuracy considering different ablations of \method{} on the study sites (results averaged over ten random splits)}
\label{tab:component}
\centering
 \begin{tabular}{|c|c|c|c|c|}
  \hline
  Site & Classifier & F1 score & Kappa & Accuracy \\   \hline
  \multirow{5}*{Reunion} & $noAtt$ & 77.66 $\pm$ 0.99 & 0.749 $\pm$ 0.011 & 77.74 $\pm$ 0.99 \\
  & $SoftMaxAtt$ & 77.32 $\pm$ 1.22 & 0.746 $\pm$ 0.013 & 77.47 $\pm$ 1.18 \\
  & $noHierPre$ & 78.35 $\pm$ 0.70 & 0.756 $\pm$ 0.007 & 78.43 $\pm$ 0.66 \\
  & $noEnrich$ & 79.09 $\pm$ 0.57 & 0.764 $\pm$ 0.006 & 79.10 $\pm$ 0.50 \\
  & \method & \textbf{79.66} $\pm$ 0.85 & \textbf{0.772} $\pm$ 0.009 & \textbf{79.78} $\pm$ 0.82 \\
  \hline
  \multirow{5}*{Senegal} & $noAtt$ & 89.86 $\pm$ 0.62 & 0.874 $\pm$ 0.008 & 89.89 $\pm$ 0.63 \\
  & $SoftMaxAtt$ & 89.91 $\pm$ 0.54 & 0.874 $\pm$ 0.007 & 89.92 $\pm$ 0.52 \\ 
  & $noHierPre$ & 89.25 $\pm$ 0.88 & 0.866 $\pm$ 0.011 & 89.24 $\pm$ 0.87 \\
  & $noEnrich$ & 89.12 $\pm$ 0.64 & 0.864 $\pm$ 0.008 & 89.11 $\pm$ 0.64 \\
  & \method & \textbf{90.78} $\pm$ 1.03 & \textbf{0.885} $\pm$ 0.013 & \textbf{90.78} $\pm$ 1.03 \\
  \hline
 \end{tabular}
\end{table}

\subsubsection{Assessment of the NDVI index helpfulness}
As additional experiment, we evaluate here if the NDVI index as additional optical descriptor has an impact on the final land cover mapping obtained using the \method{} architecture. Indeed, considering NDVI index as additional feature in land cover classification task was obvious when training conventional machine learning algorithms since such techniques cannot extract specialized features for a specific task at hand~\cite{Lecun2015}. Nowadays, \textcolor{black}{the new paradigm related to deep (or representational) learning~\cite{Lecun2015}} \textcolor{black}{is emerging and demonstrating to be more and more effective in the field of remote sensing~\cite{Ma2019}}. Neural networks have the ability to extract features optimised for a specific task (when enough data are available) \textcolor{black}{avoiding the necessity} to extract hand-crafted features. Thus, employing spectral indices like NDVI as additional features to deal with land cover classification could not be necessary when using neural networks. Therefore, we evaluate on the \textcolor{black}{two} study sites our model performances when excluding the NDVI index in the input (optical) time series. We named such variant of the model $noNDVI$. Results are reported in Table \ref{tab:ndvi}. 

\begin{table}[!htbp]
\caption{F1 score, Kappa and Accuracy considering the exclusion of NDVI index on the study sites (results averaged over ten random splits)}
\label{tab:ndvi}
\centering
 \begin{tabular}{|c|c|c|c|c|}
  \hline
  Site & Classifier & F1 score & Kappa & Accuracy \\   \hline
  \multirow{2}*{Reunion} & $noNDVI$ & \textbf{79.83} $\pm$ 0.70 & \textbf{0.774} $\pm$ 0.008 & \textbf{79.95} $\pm$ 0.68 \\
  & \method & 79.66 $\pm$ 0.85 & 0.772 $\pm$ 0.009 & 79.78 $\pm$ 0.82 \\
  \hline
  \multirow{2}*{Senegal} & $noNDVI$ & 90.46 $\pm$ 0.82 & 0.881 $\pm$ 0.010 & 90.46 $\pm$ 0.82 \\
  & \method & \textbf{90.78} $\pm$ 1.03 & \textbf{0.885} $\pm$ 0.013 & \textbf{90.78} $\pm$ 1.03 \\
  \hline
 \end{tabular}
\end{table}

We can note on both study sites that there is no significant difference between \method{} and $noNDVI$ performances. $noNDVI$ performs slightly better than \method{} on the \textit{Reunion island} and inversely on the \textit{Senegalese} site. These small variations can come from model properties such as kernel weight initialization or parameters optimization that can induce such kind of performance fluctuations.

To conclude, this experiment underlines that our model, considering the two study sites involved in the experimental evaluation, is able to overcome the use of such common hand-crafted features achieving the same performances in the land cover mapping task. \textcolor{black}{Such result makes a step further on the comprehension of which hand-crafted features are convenient (or not) to be extracted during the preprocessing step as well as save time, computation and storage resources during the analysis pipeline.}

\subsection{Qualitative analysis of land cover maps}
\begin{figure}[!htbp]
\centering
\begin{tabular}{ccc}
 \multicolumn{3}{l}{Detail 1: A mixed coastal urban and agricultural area} \\
 \subfloat[\label{fig:ref-reunion1}RGB Image]{\includegraphics[width=0.33\textwidth]{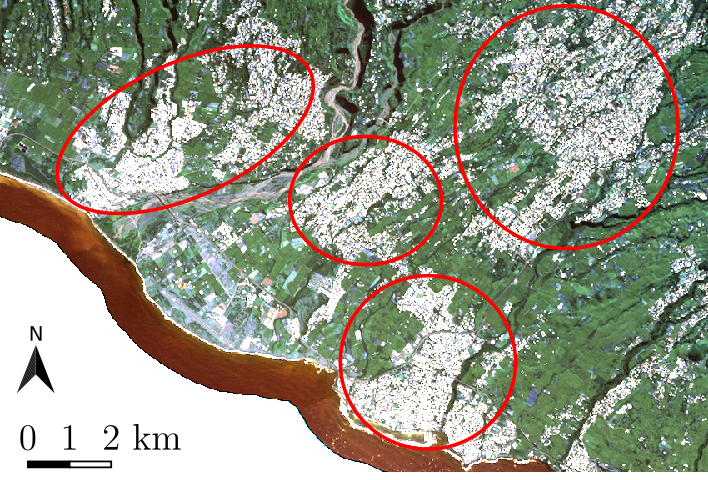}} & 
 \subfloat[\label{fig:rf-reunion1}RF]{\includegraphics[width=0.33\textwidth]{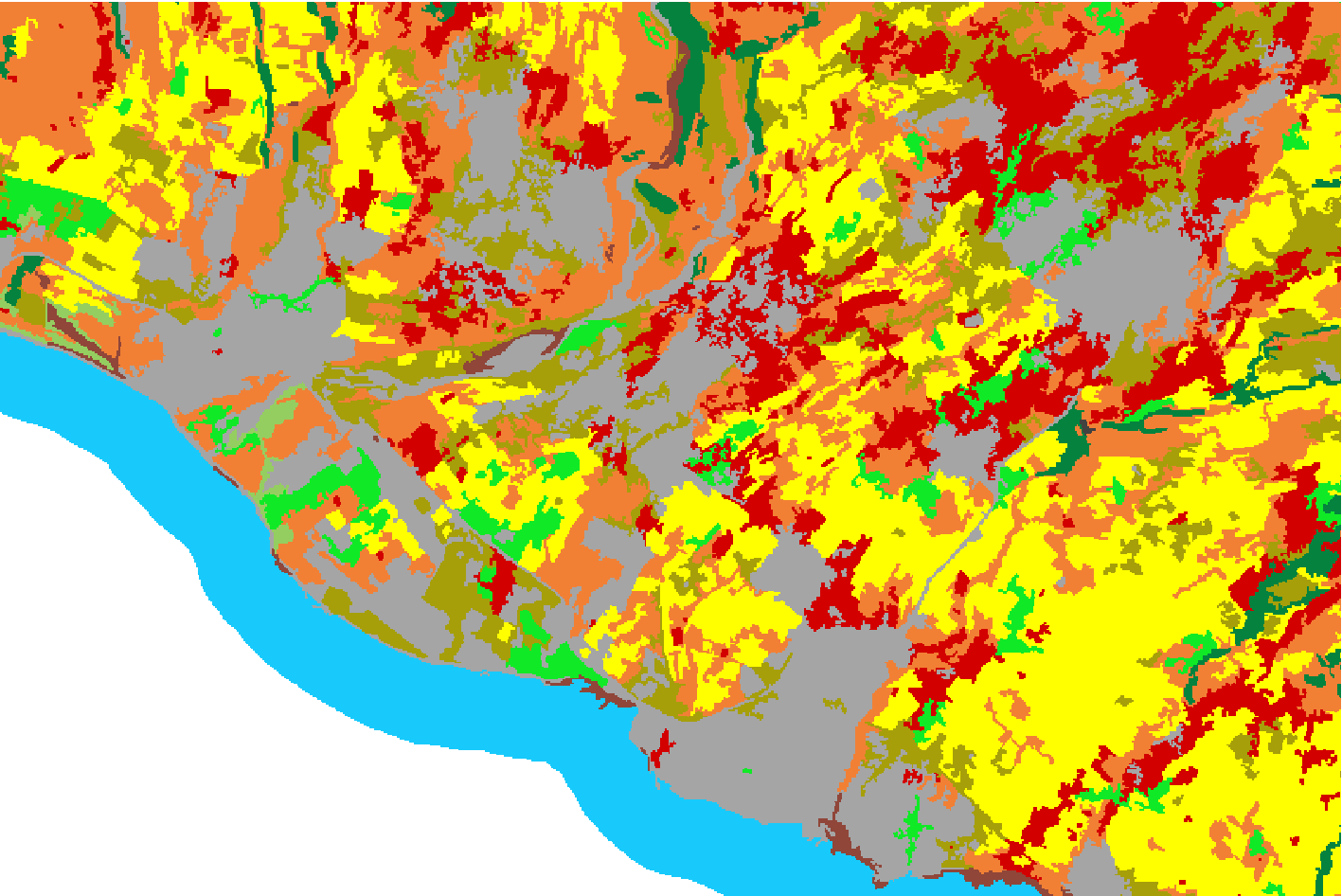}} & 
 \subfloat[\label{fig:svm-reunion1}SVM]{\includegraphics[width=0.33\textwidth]{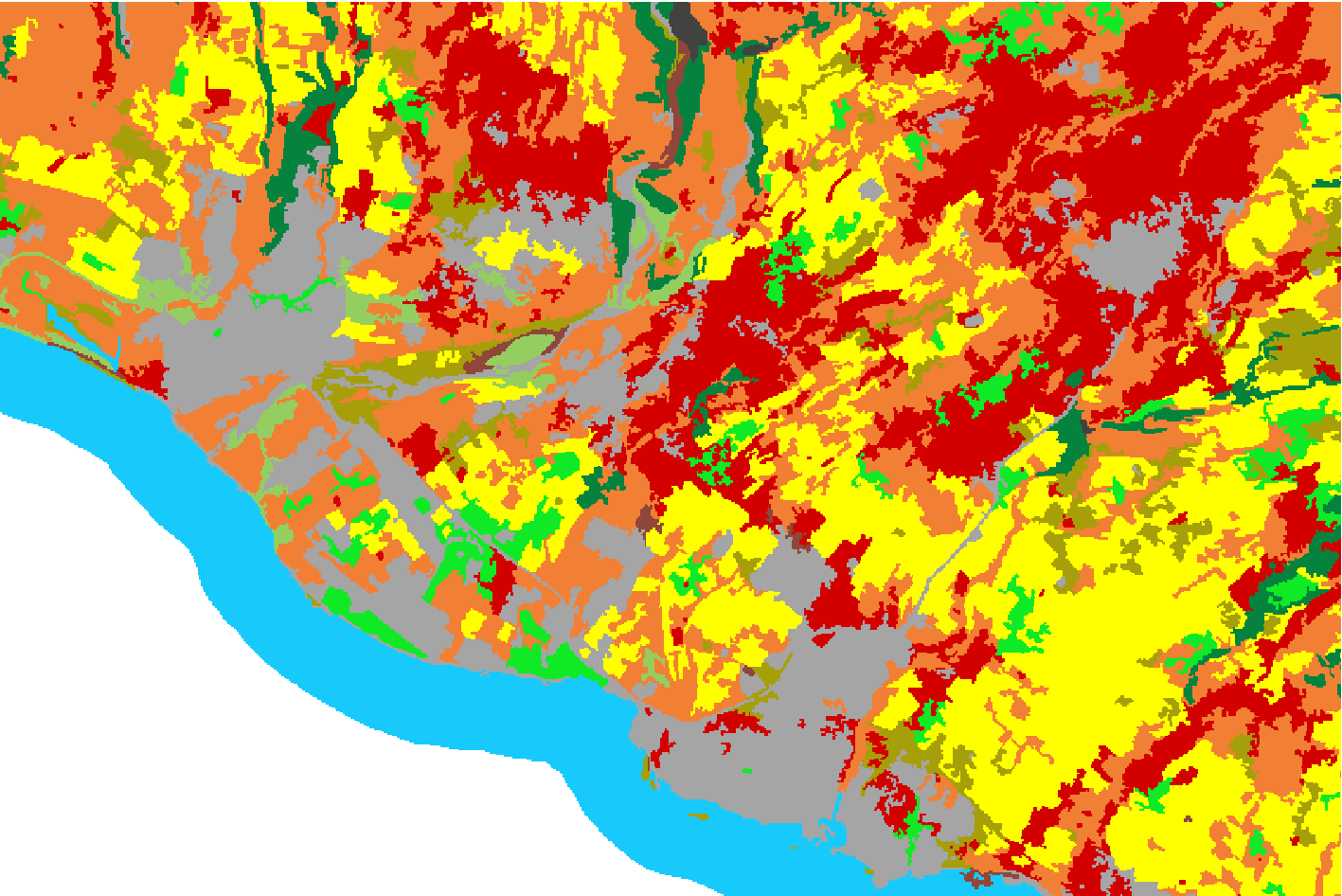}} \\
 \subfloat[\label{fig:mlp-reunion1}MLP]{\includegraphics[width=0.33\textwidth]{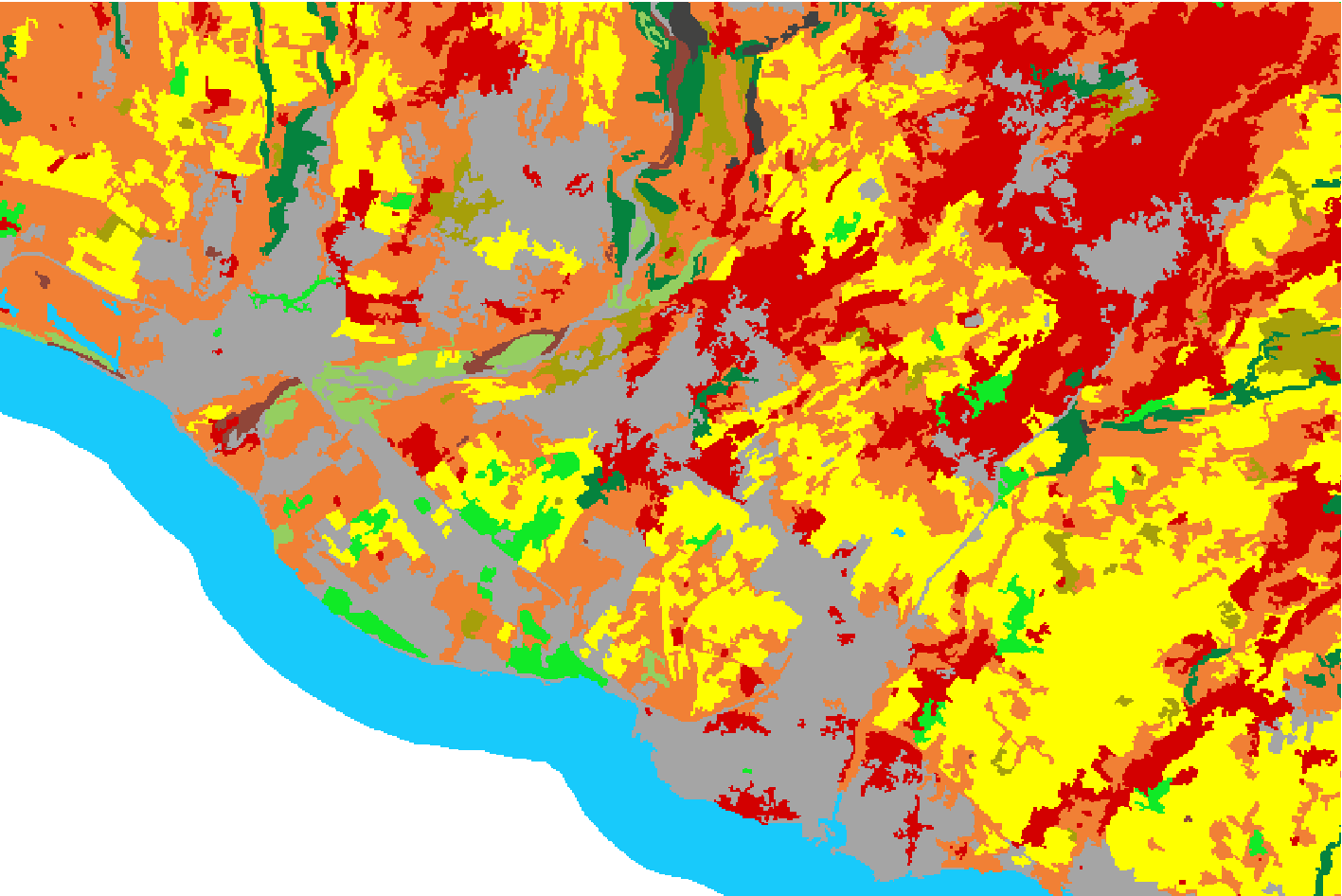}} & 
 \subfloat[\label{fig:model-reunion1}\method{}]{\includegraphics[width=0.33\textwidth]{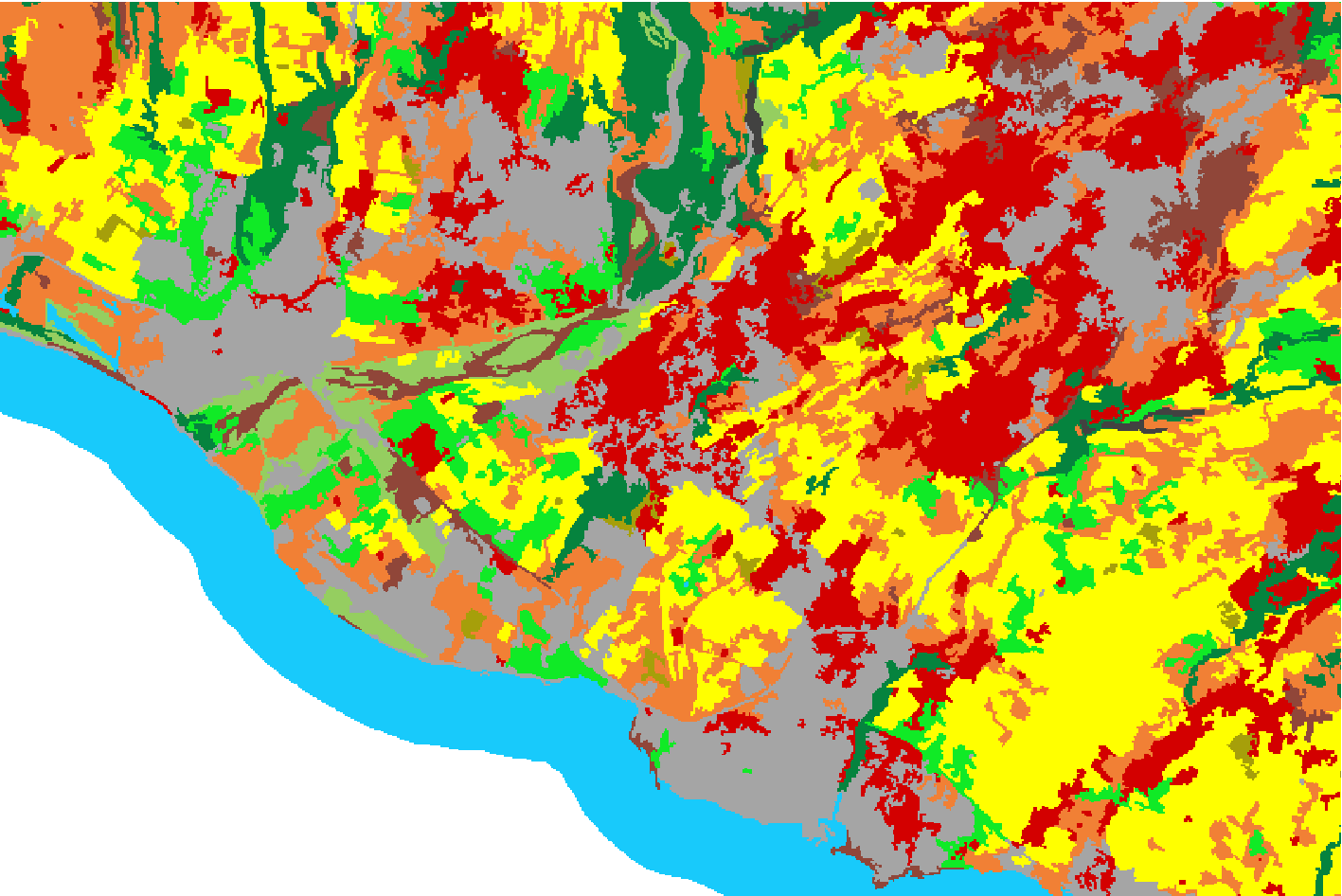}}
\\
\multicolumn{3}{l}{Detail 2: A mixed agricultural and natural vegetation area} \\
 \subfloat[\label{fig:ref-reunion2}RGB Image]{\includegraphics[width=0.33\textwidth]{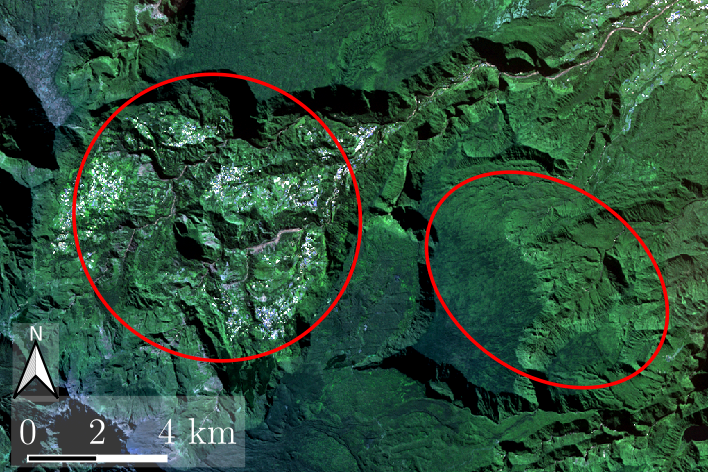}} & 
 \subfloat[\label{fig:rf-reunion2}RF]{\includegraphics[width=0.33\textwidth]{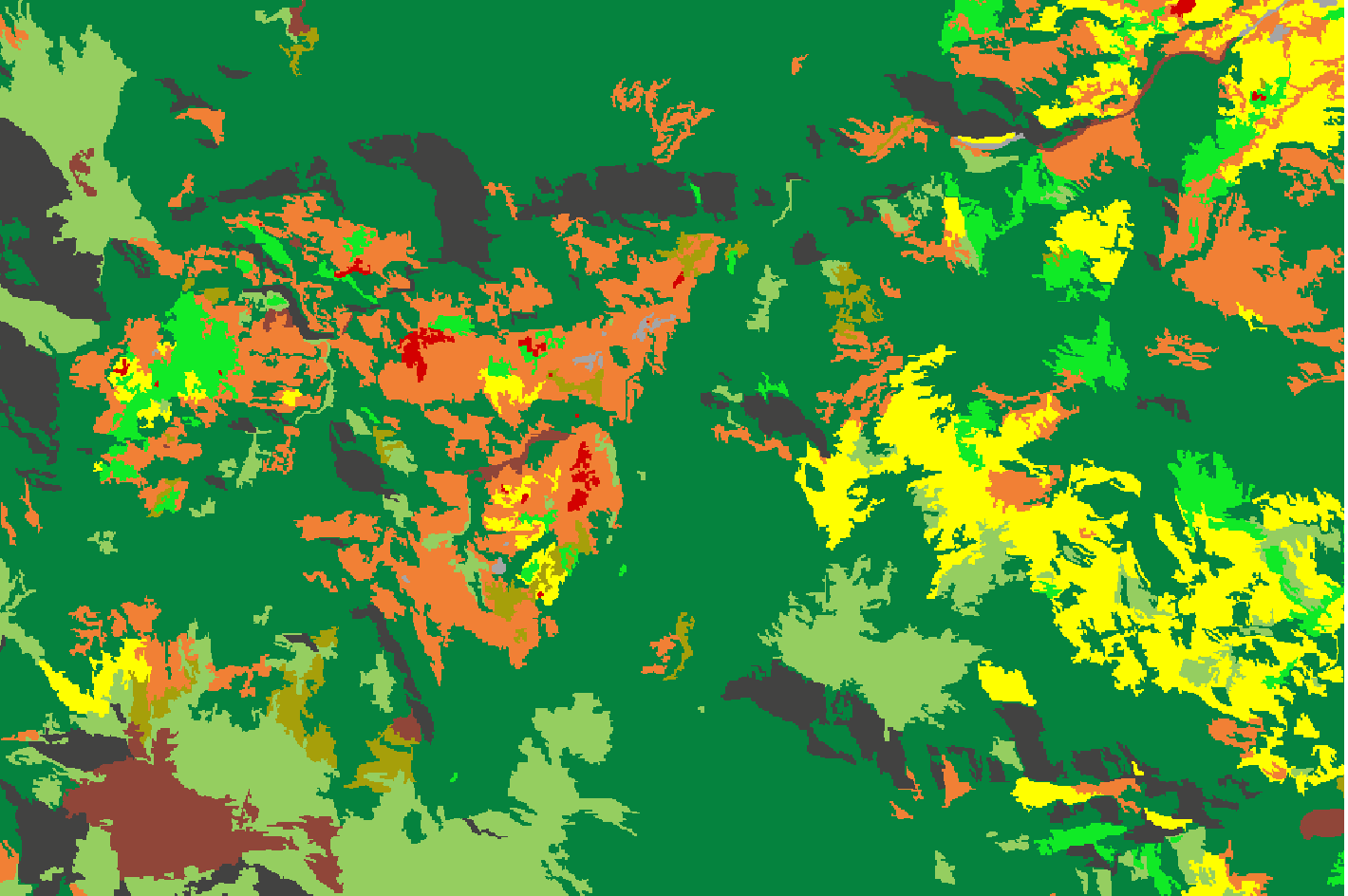}} & 
 \subfloat[\label{fig:svm-reunion2}SVM]{\includegraphics[width=0.33\textwidth]{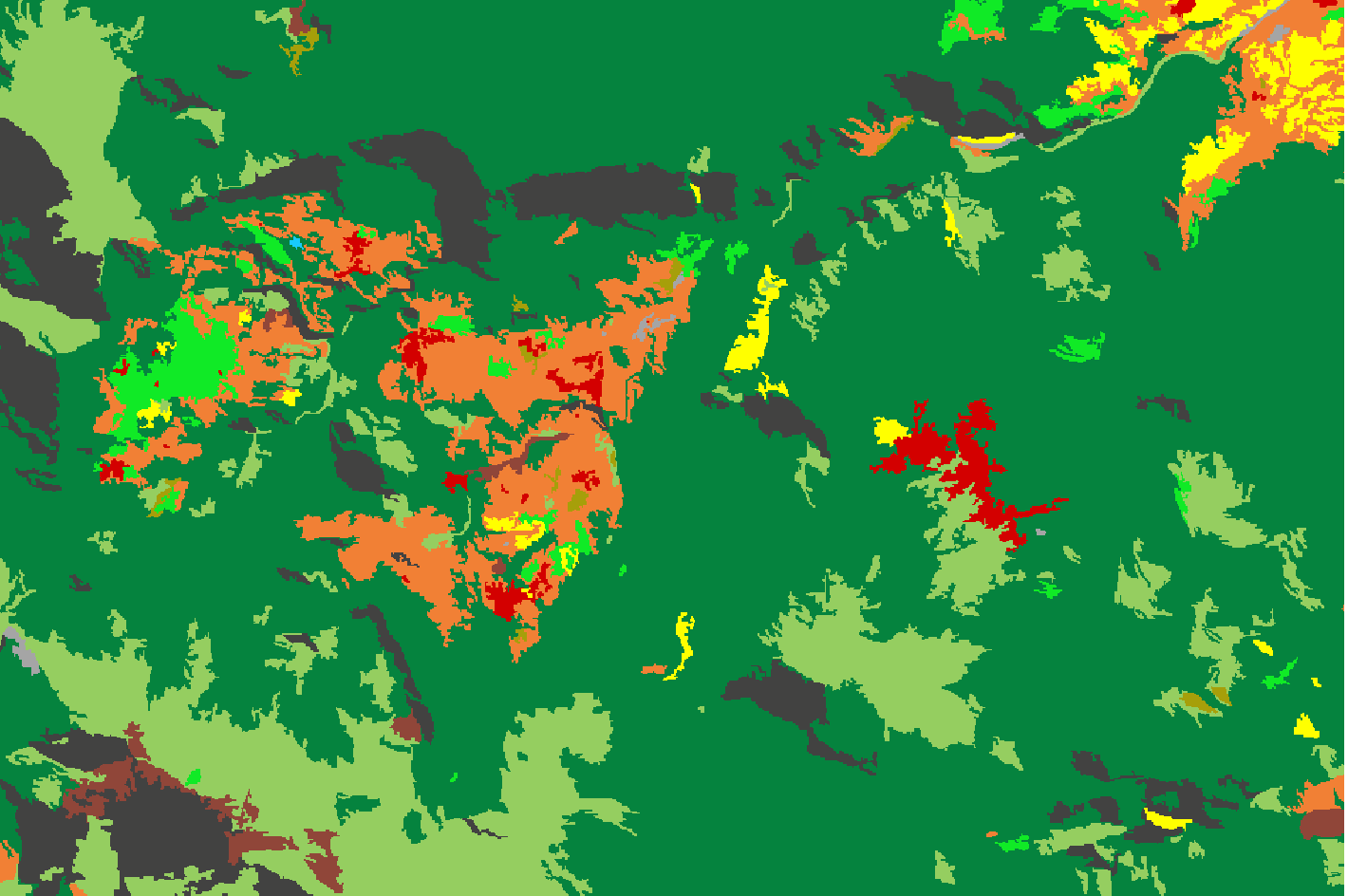}} \\
 \subfloat[\label{fig:mlp-reunion2}MLP]{\includegraphics[width=0.33\textwidth]{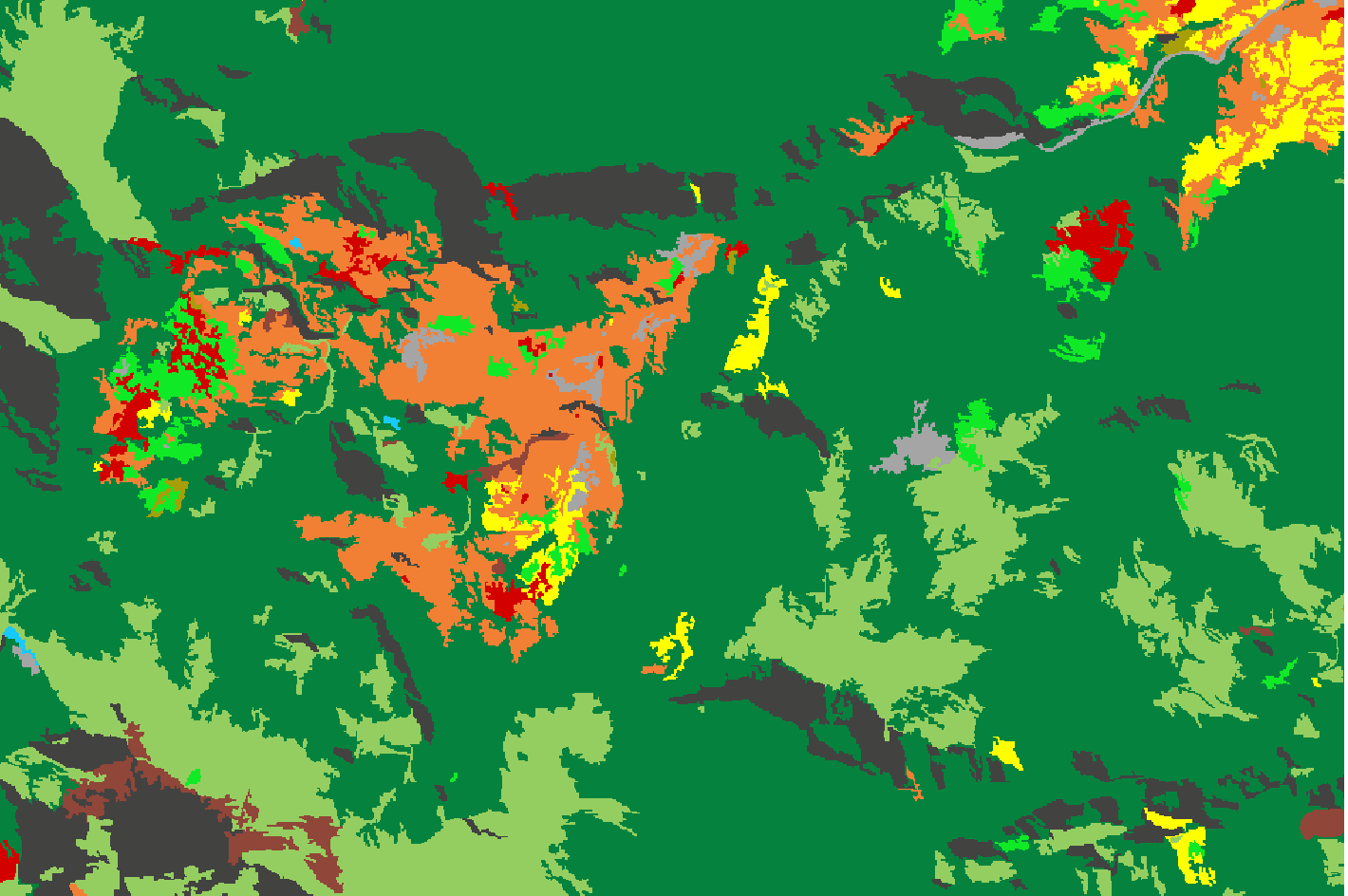}} & 
 \subfloat[\label{fig:model-reunion2}\method{}]{\includegraphics[width=0.33\textwidth]{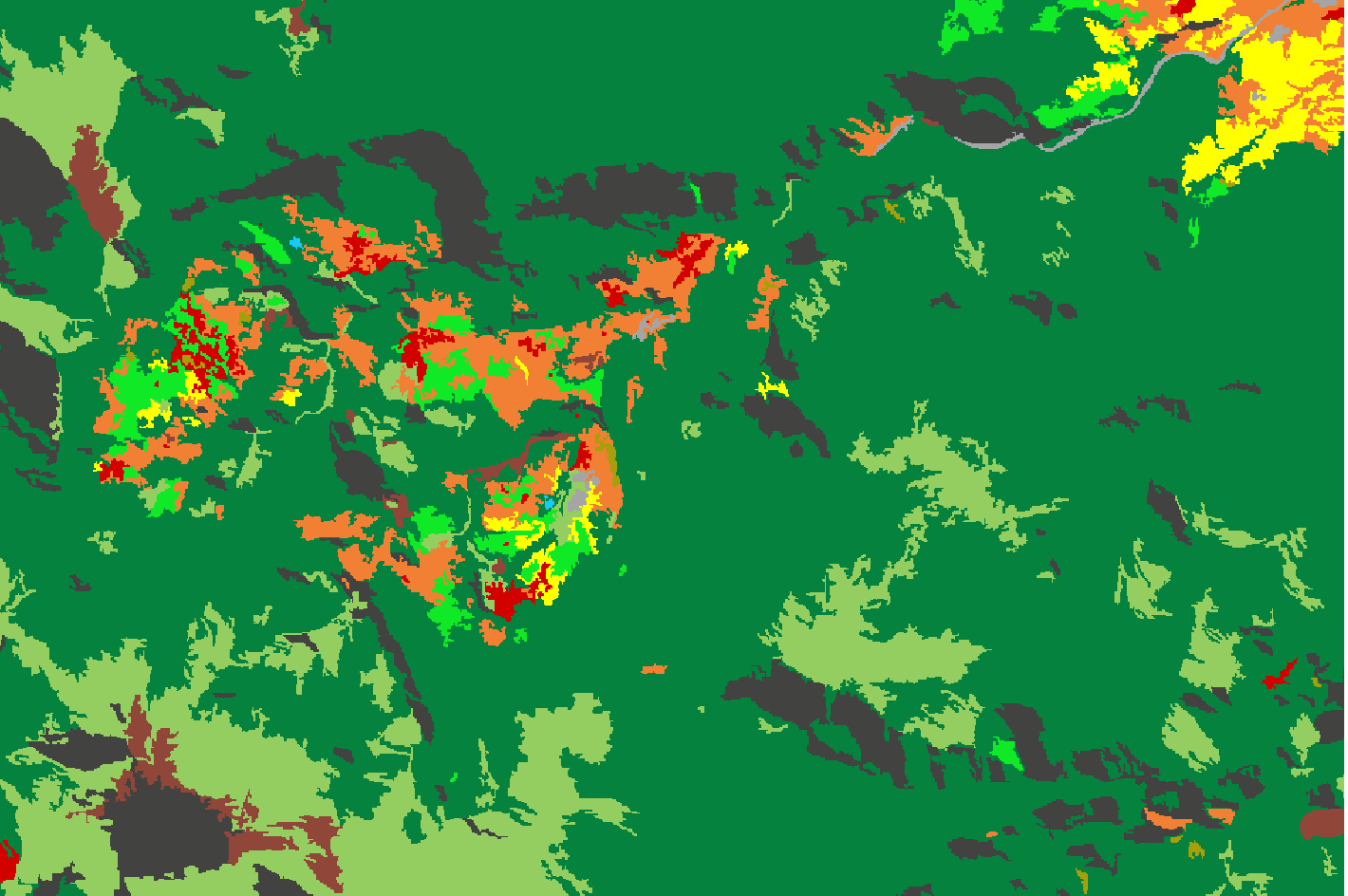}}
 \\
 \multicolumn{3}{c}{\includegraphics[width=\textwidth]{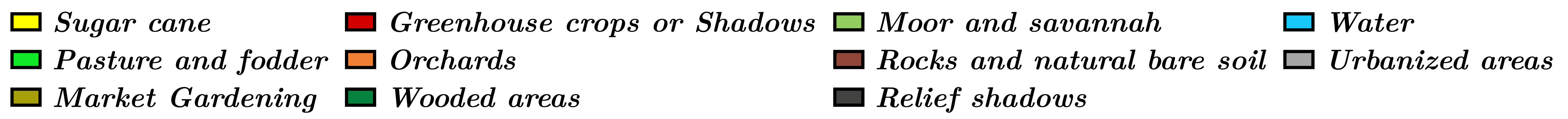}}

\end{tabular}

\caption{Qualitative investigation of land cover map details produced on the
\textit{Reunion island} study site by RF, SVM, MLP and \method{} on mixed urban/agricultural (top) and agricultural/natural vegetation (bottom) areas.}
\label{fig:map-reunion}
\end{figure}

\begin{figure}[!htbp]
\centering
\begin{tabular}{ccc}
 \multicolumn{3}{l}{Detail 1: A rural landscape including a wet area} \\
 \subfloat[\label{fig:ref-senegal1}RGB Image]{\includegraphics[width=0.33\textwidth]{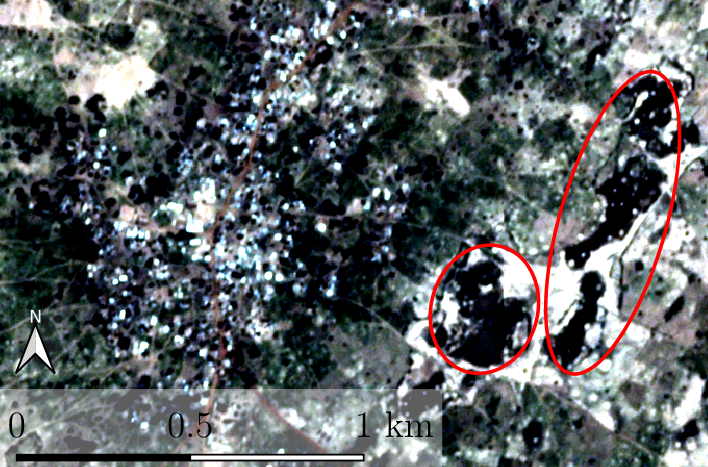}} & 
 \subfloat[\label{fig:rf-senegal1}RF]{\includegraphics[width=0.33\textwidth]{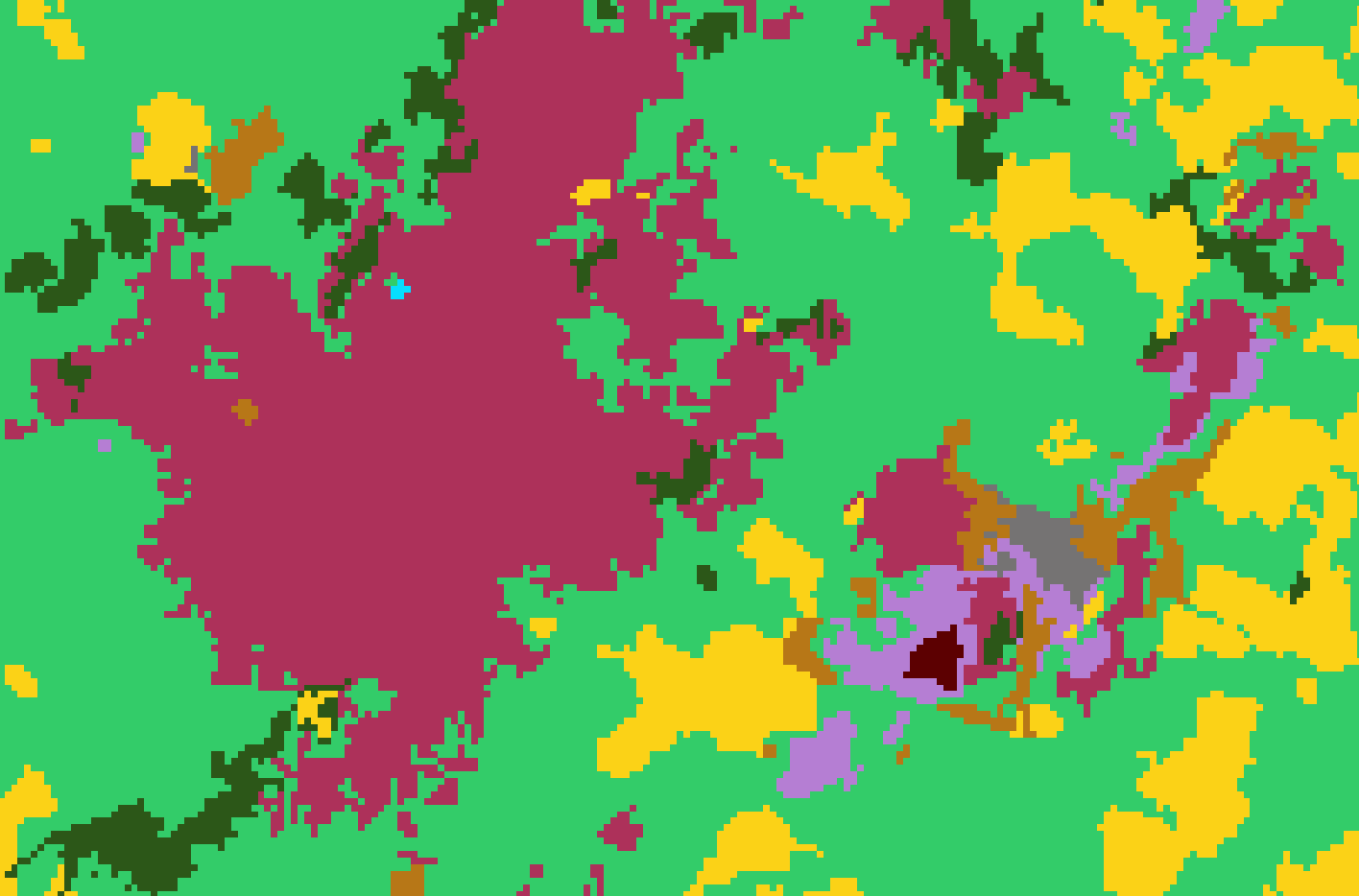}} & 
 \subfloat[\label{fig:svm-senegal1}SVM]{\includegraphics[width=0.33\textwidth]{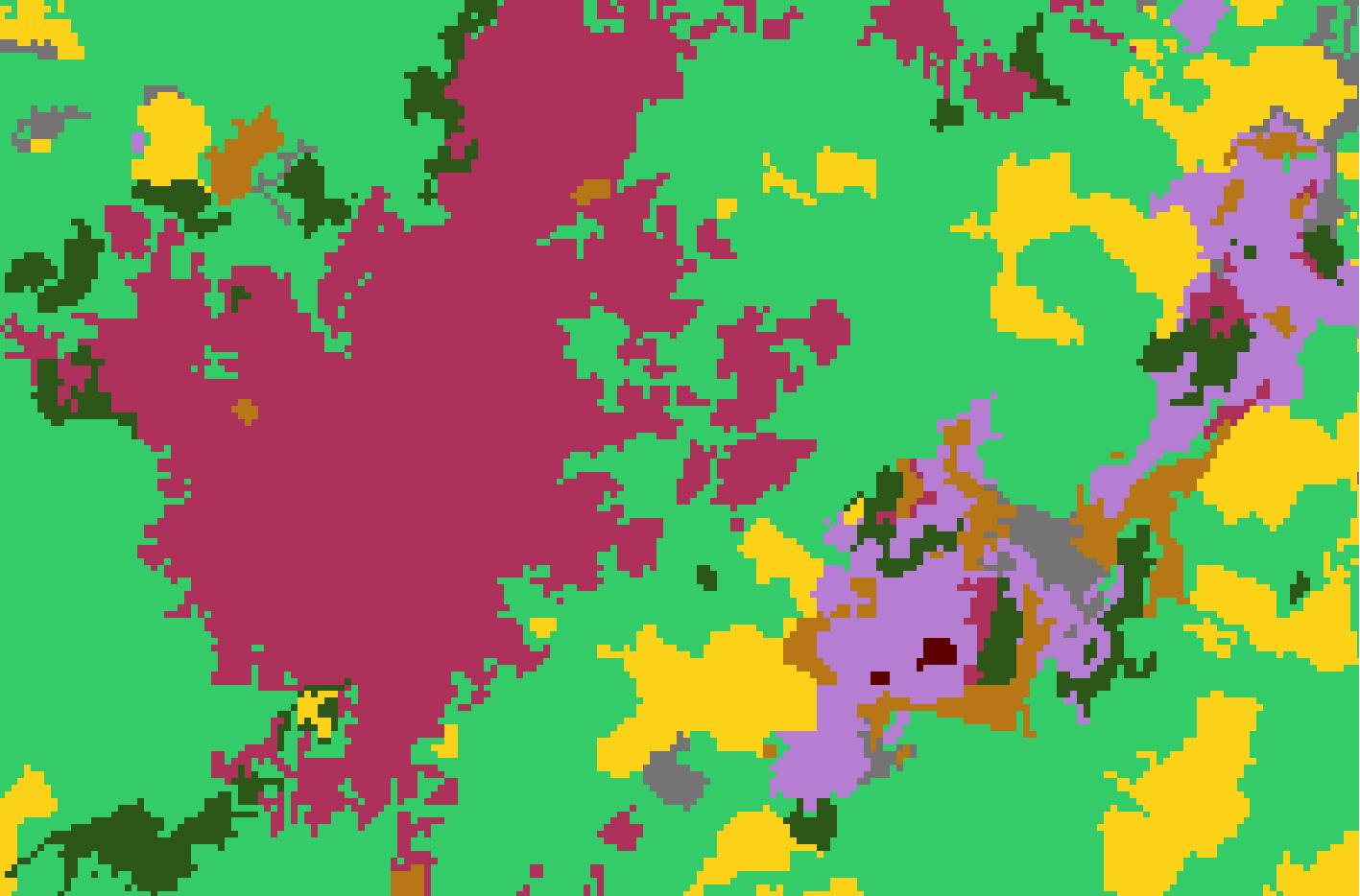}} \\
 \subfloat[\label{fig:mlp-senegal1}MLP]{\includegraphics[width=0.33\textwidth]{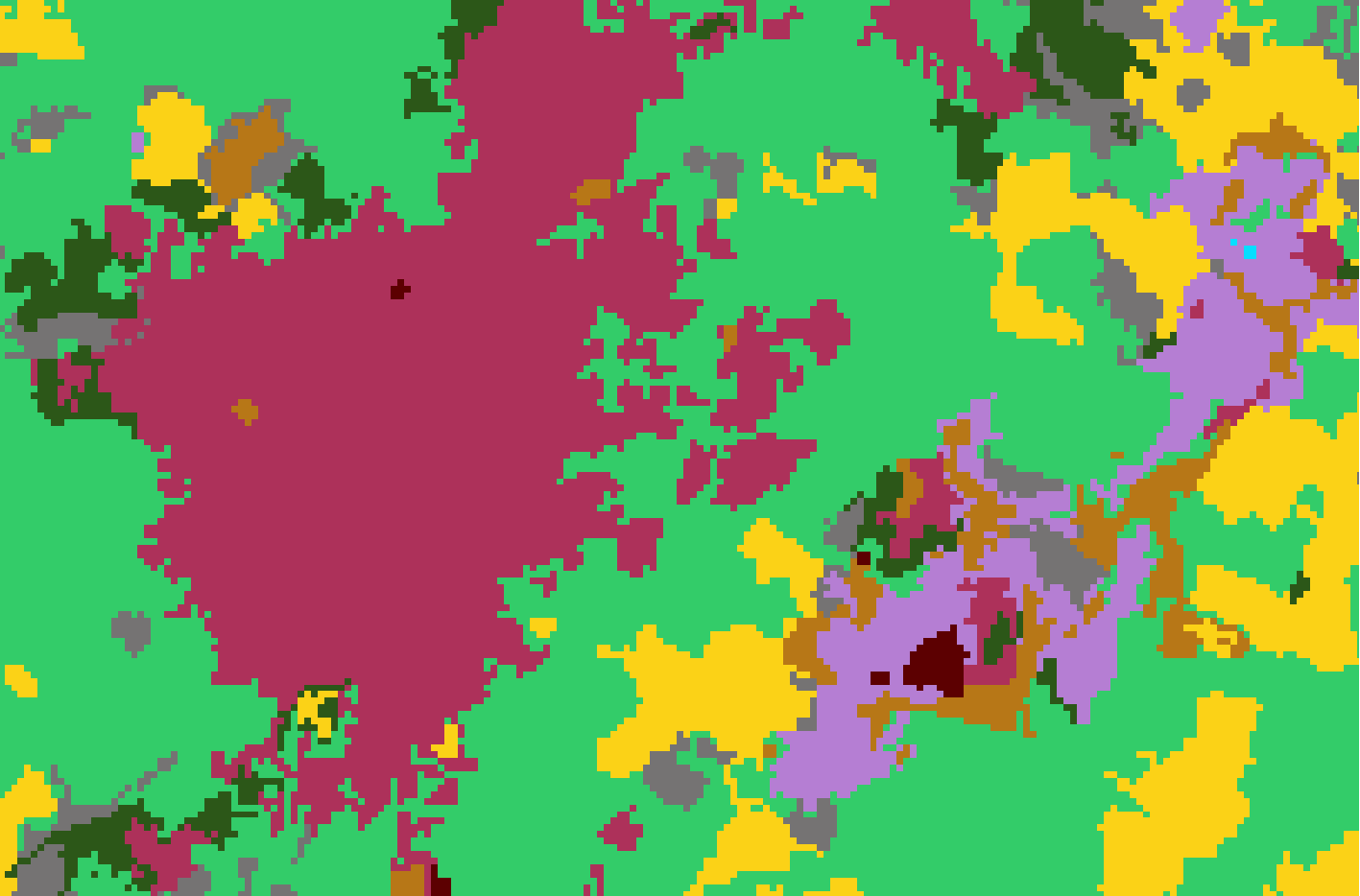}} & 
 \subfloat[\label{fig:model-senegal1}\method{}]{\includegraphics[width=0.33\textwidth]{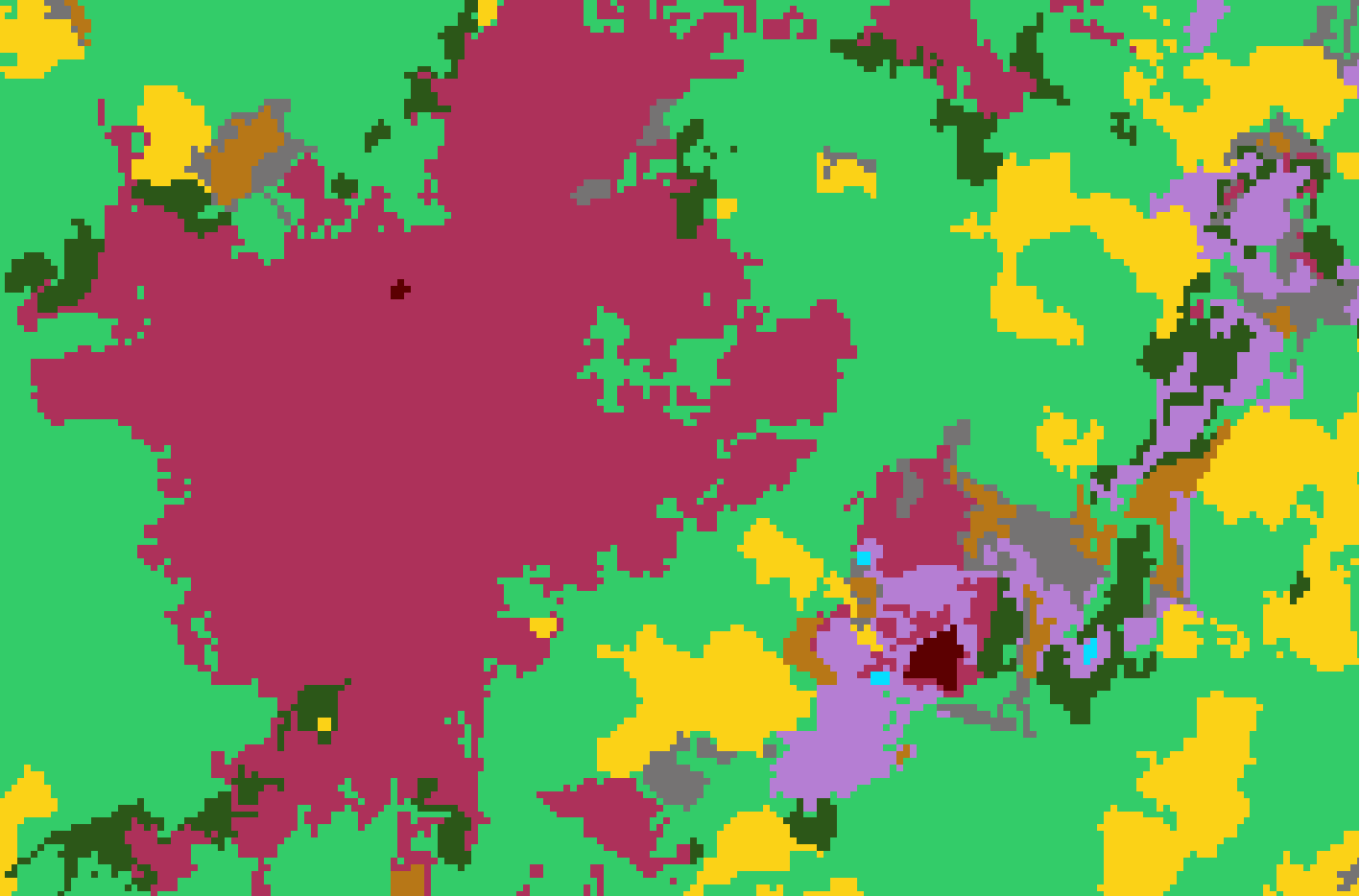}}
\\
\multicolumn{3}{l}{Detail 2: A rural landscape including buildings and agricultural activities} \\
 \subfloat[\label{fig:ref-senegal2}RGB Image]{\includegraphics[width=0.33\textwidth]{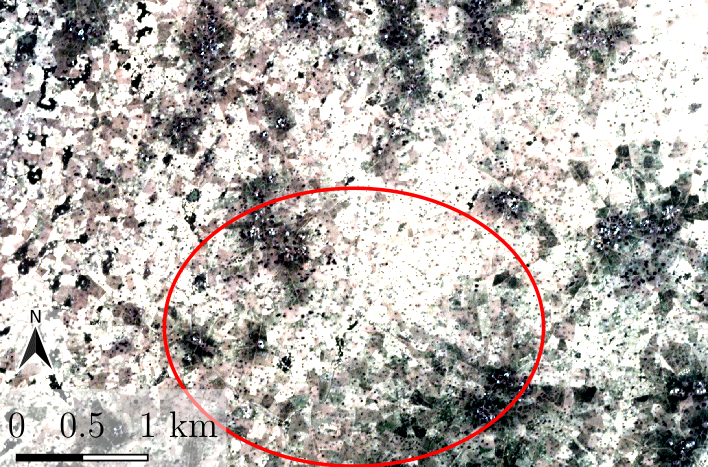}} & 
 \subfloat[\label{fig:rf-senegal2}RF]{\includegraphics[width=0.33\textwidth]{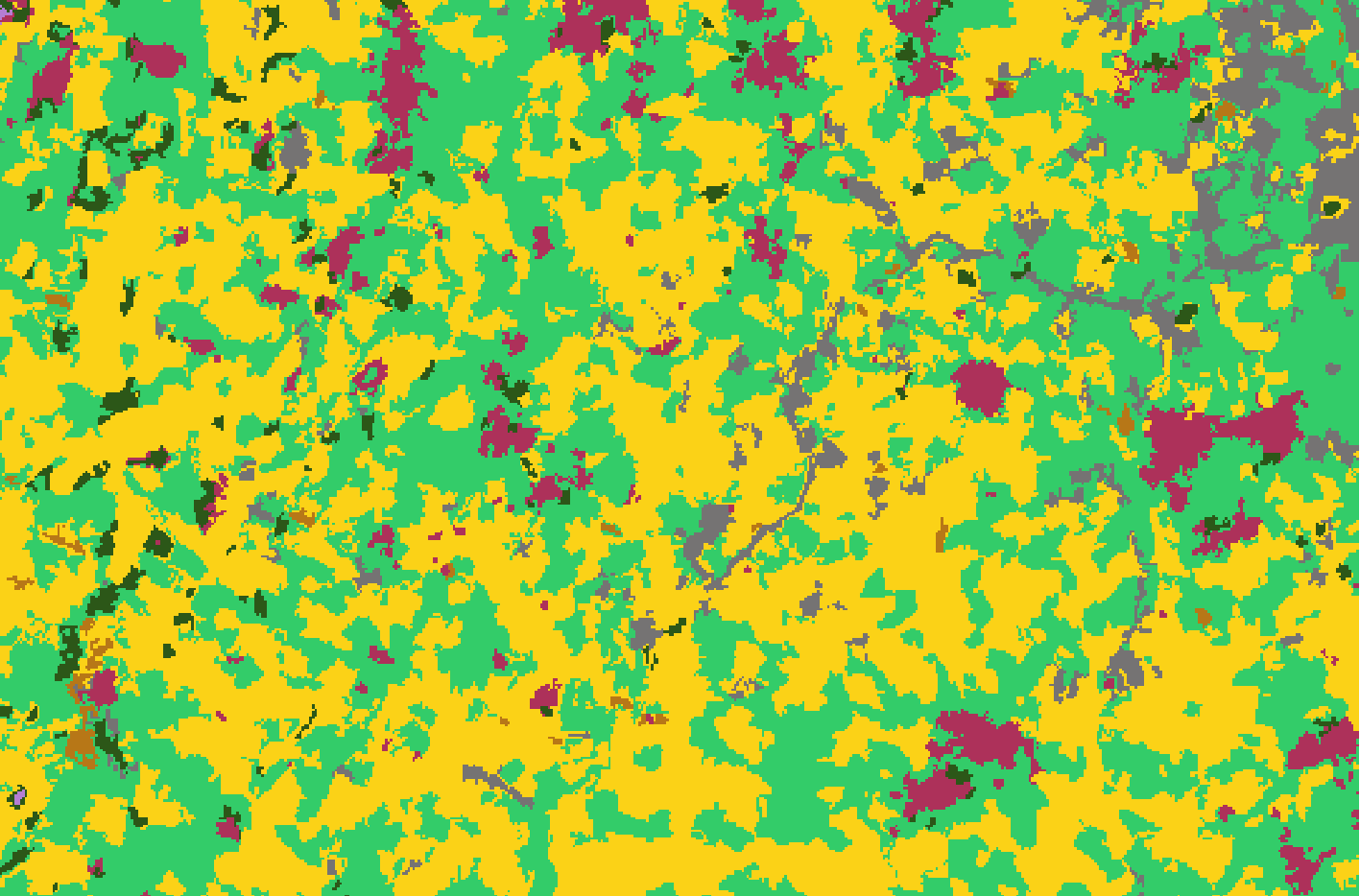}} & 
 \subfloat[\label{fig:svm-senegal2}SVM]{\includegraphics[width=0.33\textwidth]{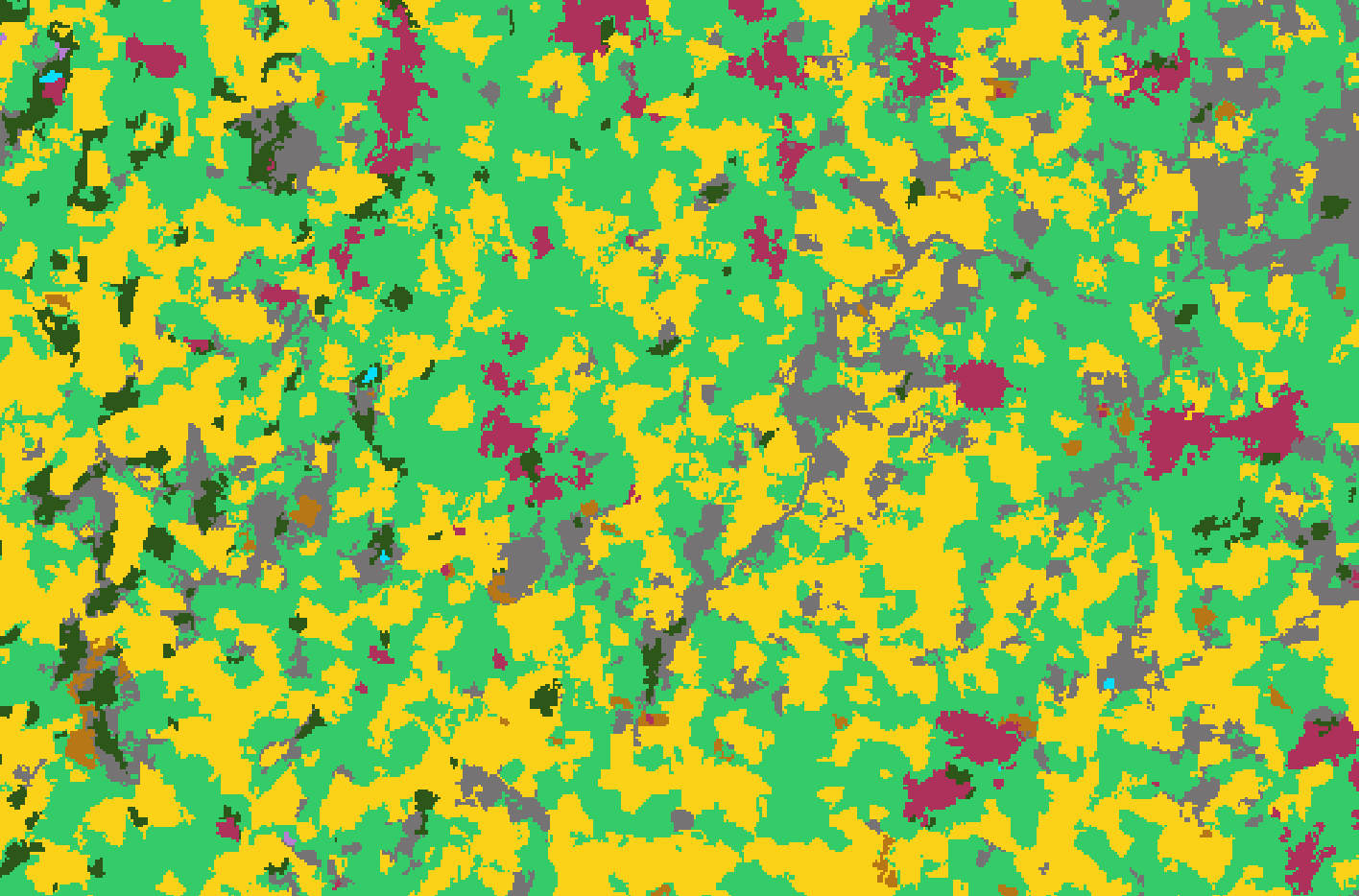}} \\
 \subfloat[\label{fig:mlp-senegal2}MLP]{\includegraphics[width=0.33\textwidth]{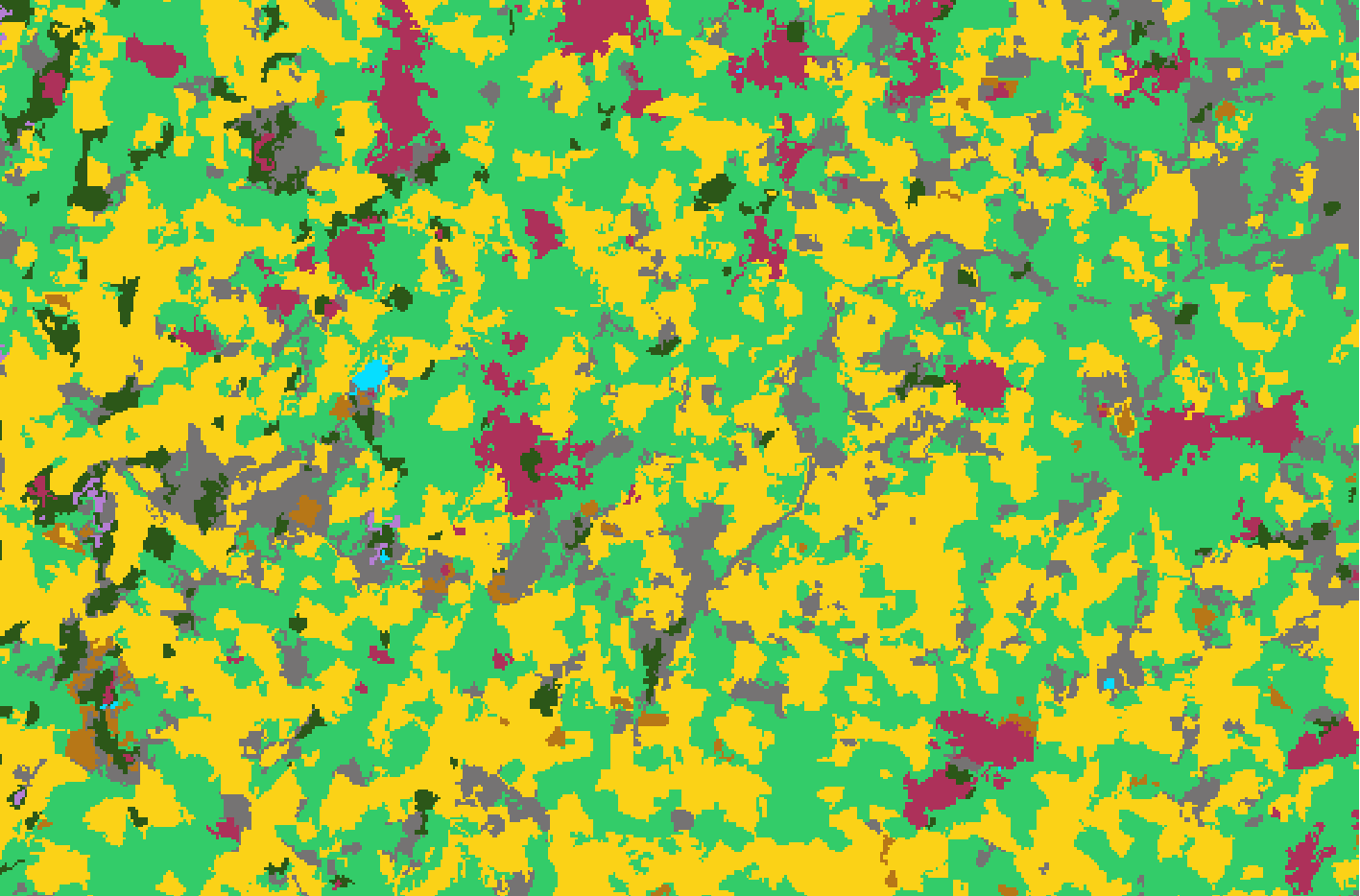}} & 
 \subfloat[\label{fig:model-senegal2}\method{}]{\includegraphics[width=0.33\textwidth]{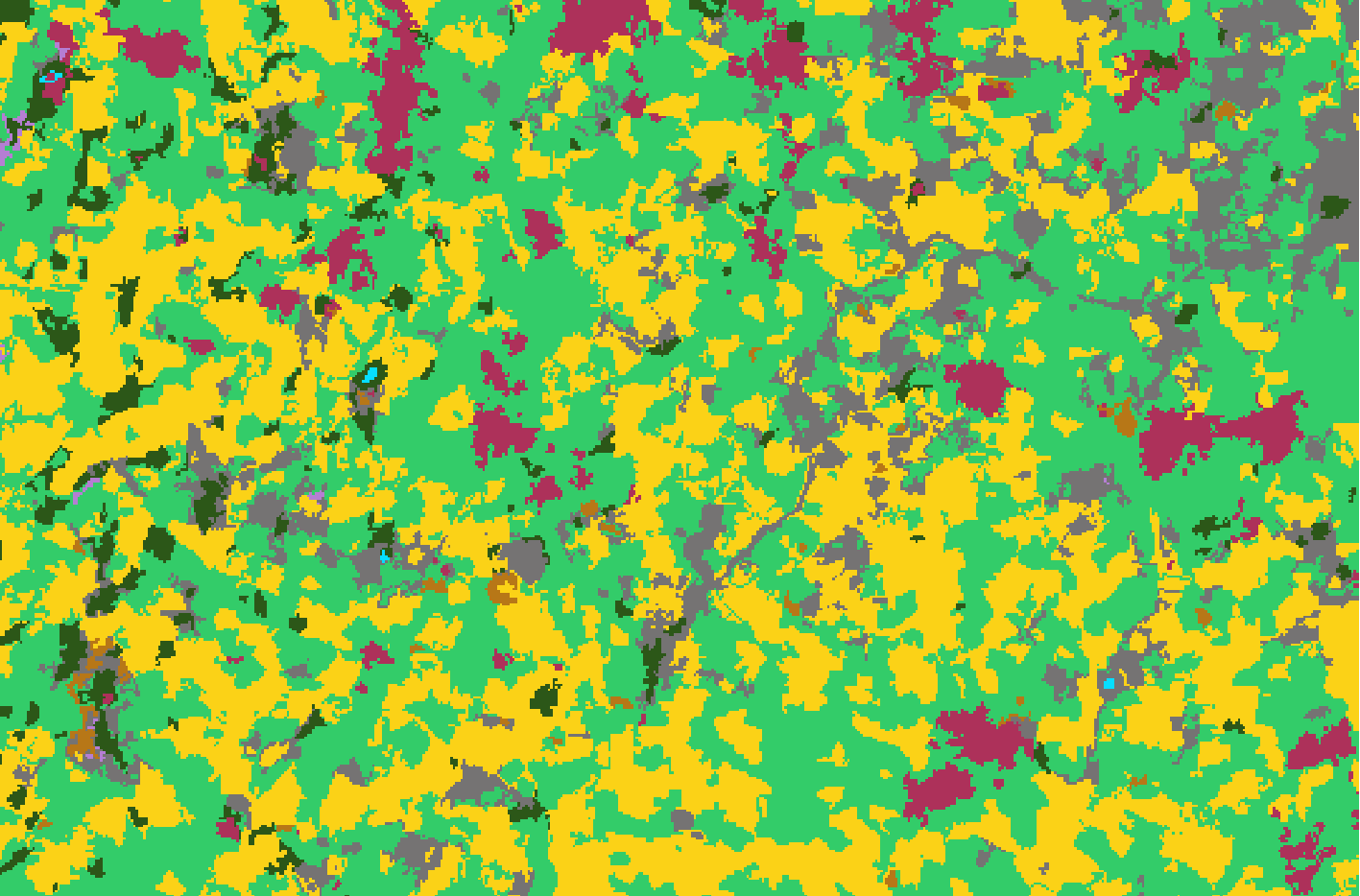}}
 \\
\multicolumn{3}{c}{\includegraphics[width=\textwidth]{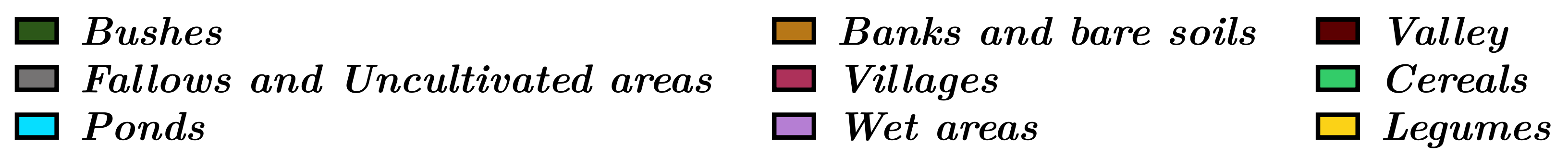}}
\end{tabular}

\caption{Qualitative investigation of land cover map details produced on the
Senegalese study site by RF, SVM, MLP and \method{} on heterogeneous landscapes including buildings, agricultural and wet areas.}
\label{fig:map-senegal}
\end{figure}

With the purpose to investigate some differences in the land cover maps produced by the competing methods (RF, SVM, MLP and \method), we highlight in Figures~\ref{fig:map-reunion} and~\ref{fig:map-senegal} some representative map details of the \textit{Reunion island} and the \textit{Senegalese} site, respectively. For each study site, we remind that land cover maps were produced by labeling each of the segments (14\,465 on the \textit{Reunion island} or 116\,937 on the \textit{Senegalese} site) obtained after the VHSR (SPOT67 or PlanetScope) image segmentation. For each map detail, we supply the corresponding VHSR image displayed in RGB colors as reference. 

Concerning \textit{Reunion island} (Figure~\ref{fig:map-reunion}), we focused in the first example (Figures~\ref{fig:rf-reunion1}, \ref{fig:svm-reunion1}, \ref{fig:mlp-reunion1} and \ref{fig:model-reunion1}) on the Saint-Pierre mixed coastal urban and agricultural area. In this example, we can note the confusions highlighted in the per-class analysis between urbanized areas and blackhouse crops. Visually, RF better classifies urbanized areas. The second example (Figures~\ref{fig:rf-reunion2}, \ref{fig:svm-reunion2}, \ref{fig:mlp-reunion2} and \ref{fig:model-reunion2}) depicts a mixed agricultural area with natural vegetation neighboring. We can note here that \method{} is the only one which detects a realistic amount of orchard cultures according to field experts. In addition, we can also observe on the right \textcolor{black}{of this extract} that RF wrongly detects sugar cane plantations instead of wooded areas, moor and savannah. 

Regarding the \textit{Senegalese} site (Figure~\ref{fig:map-senegal}), the first example (Figures \ref{fig:rf-senegal1}, \ref{fig:svm-senegal1}, \ref{fig:mlp-senegal1} and \ref{fig:model-senegal1}) depicts a wet area near the Diohine village (in the east). While SVM, MLP and \method{} tend to provide the correct representation of the wet area, RF wrongly detects villages. As pointed out in previous map details concerning Sugar cane and Orchards on the \textit{Reunion island} study site, RF predictions is sometimes biased towards most represented classes in the training data i.e. Sugar cane, Orchards in the case of \textit{Reunion island} and Villages here. In fact, RF is known to be sensible to class imbalance~\cite{Maxwell2018}. The second example focus on a rural landscape including buildings (villages) and agricultural activities. Here, RF maps much more Legumes than its competitors while the latter detect Fallows and Cereals instead. 

To sum up, these visual inspections of land cover maps are consistent with the quantitative results \textcolor{black}{previously} obtained. 

\subsection{Attention parameters analysis}
\begin{figure}[!htbp]
\centering
\centering
\subfloat[Radar branch\label{fig:att-cereals1}]{\includegraphics[width=.9\linewidth]{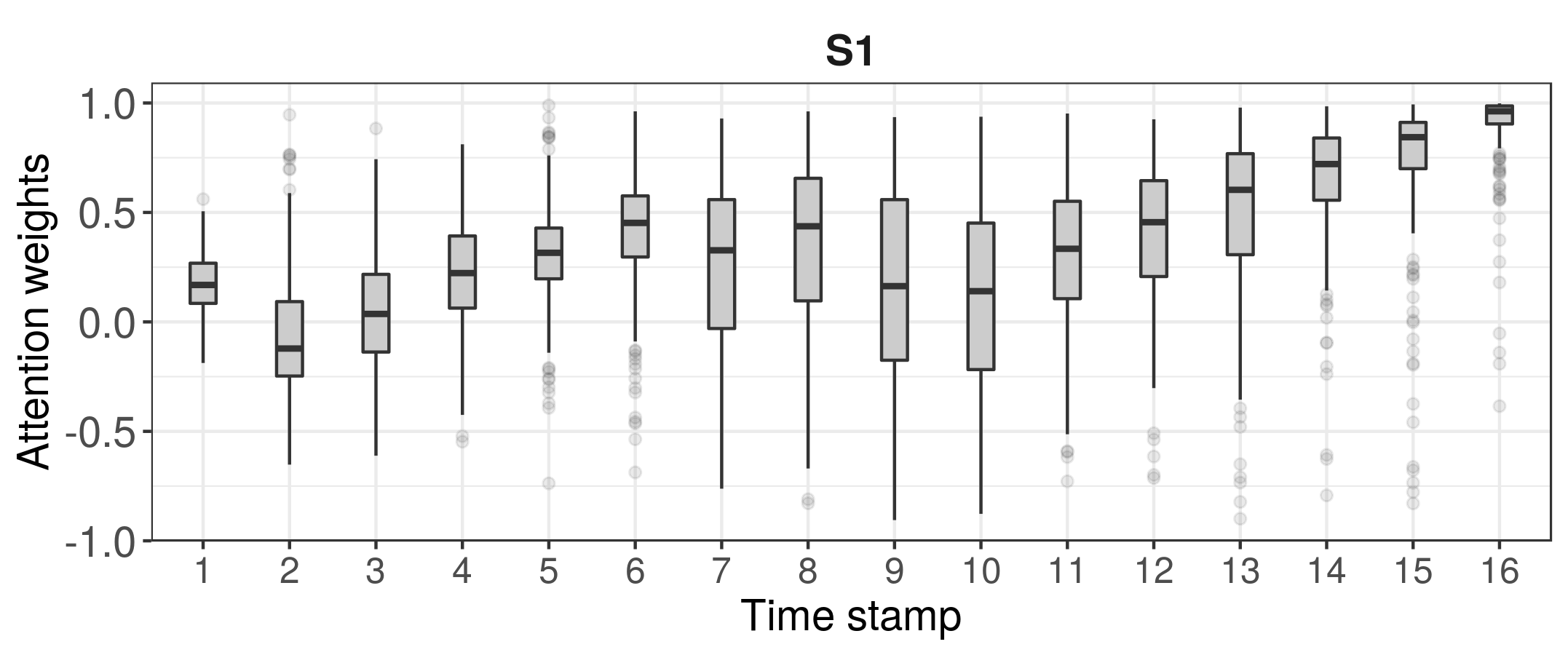}}
\hfill
\subfloat[Optical branch\label{fig:att-cereals2}]{\includegraphics[width=.9\linewidth]{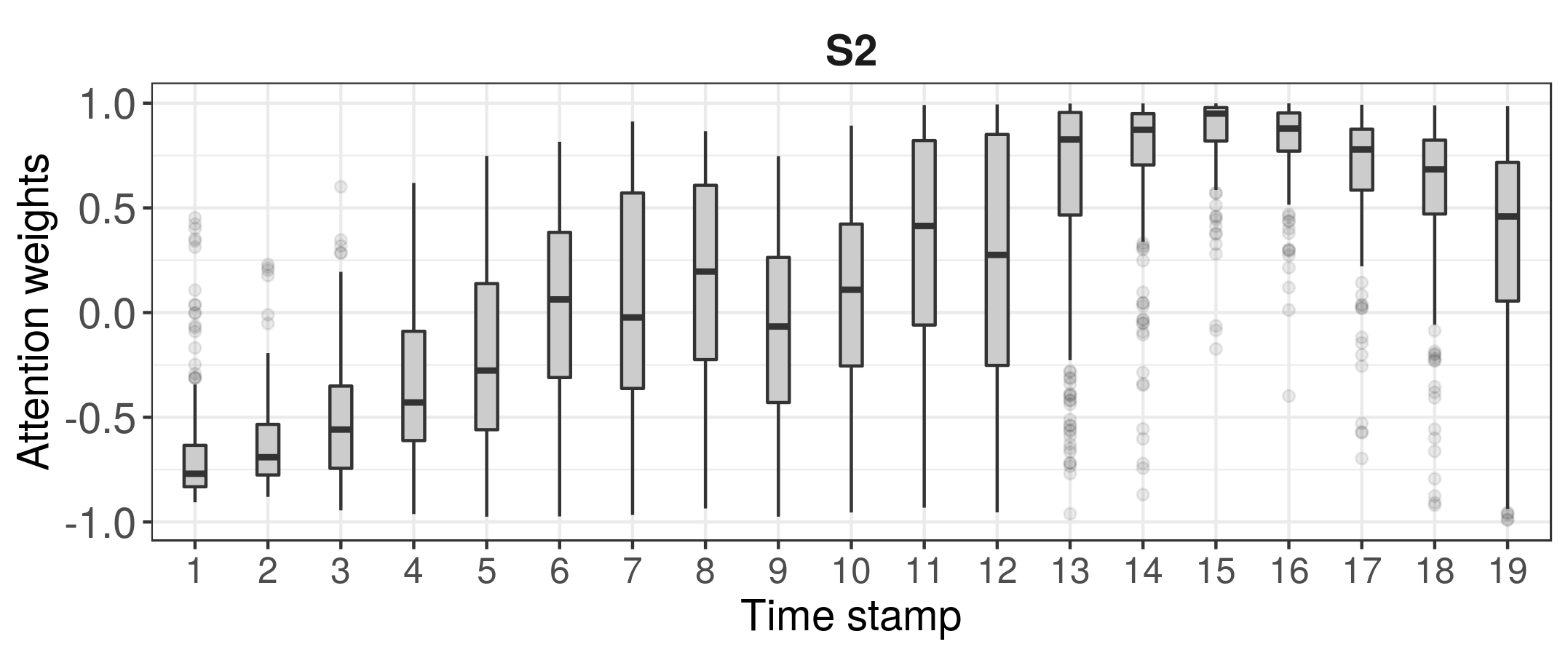}}
\hfill
\subfloat[Concatenated sources\label{fig:att-cereals3}]{\includegraphics[width=.9\linewidth]{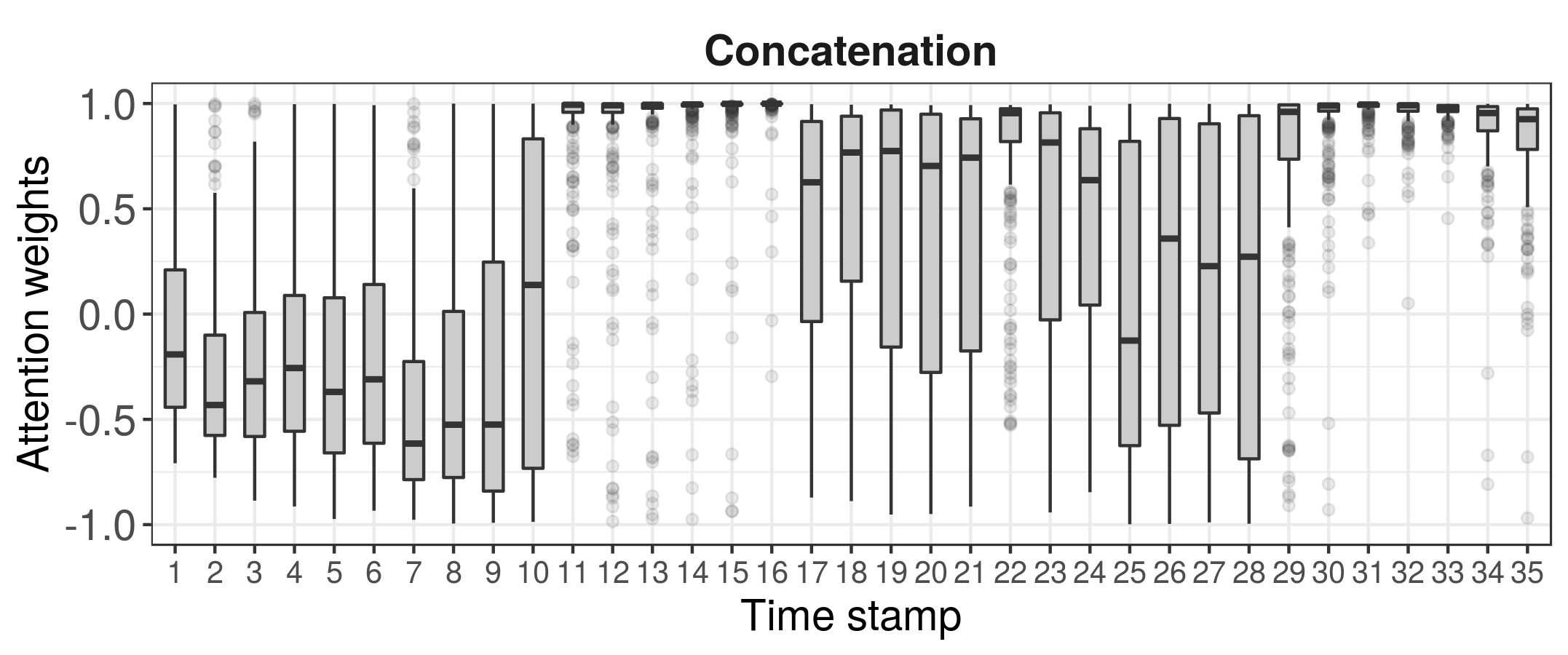}}
\caption{Box plots of the Cereals class attention weights for the radar (S1) SITS (16 time stamps), the optical (S2) SITS (19 time stamps) and the concatenated sources (radar followed by optical).}
\label{fig:att-cereals}
\end{figure}

\begin{figure}[!htbp]
\centering
\centering
\subfloat[Radar branch\label{fig:att-legumes1}]{\includegraphics[width=.9\linewidth]{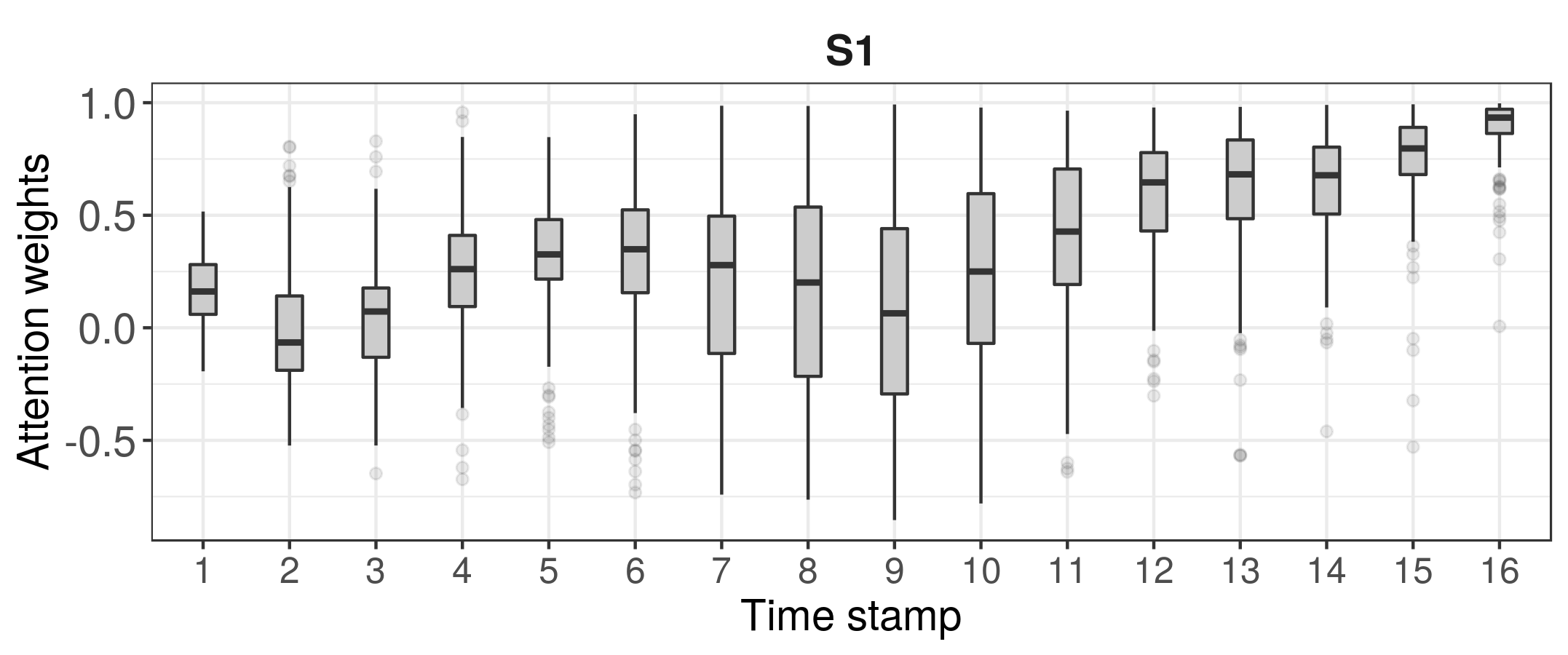}}
\hfill
\subfloat[Optical branch\label{fig:att-legumes2}]{\includegraphics[width=.9\linewidth]{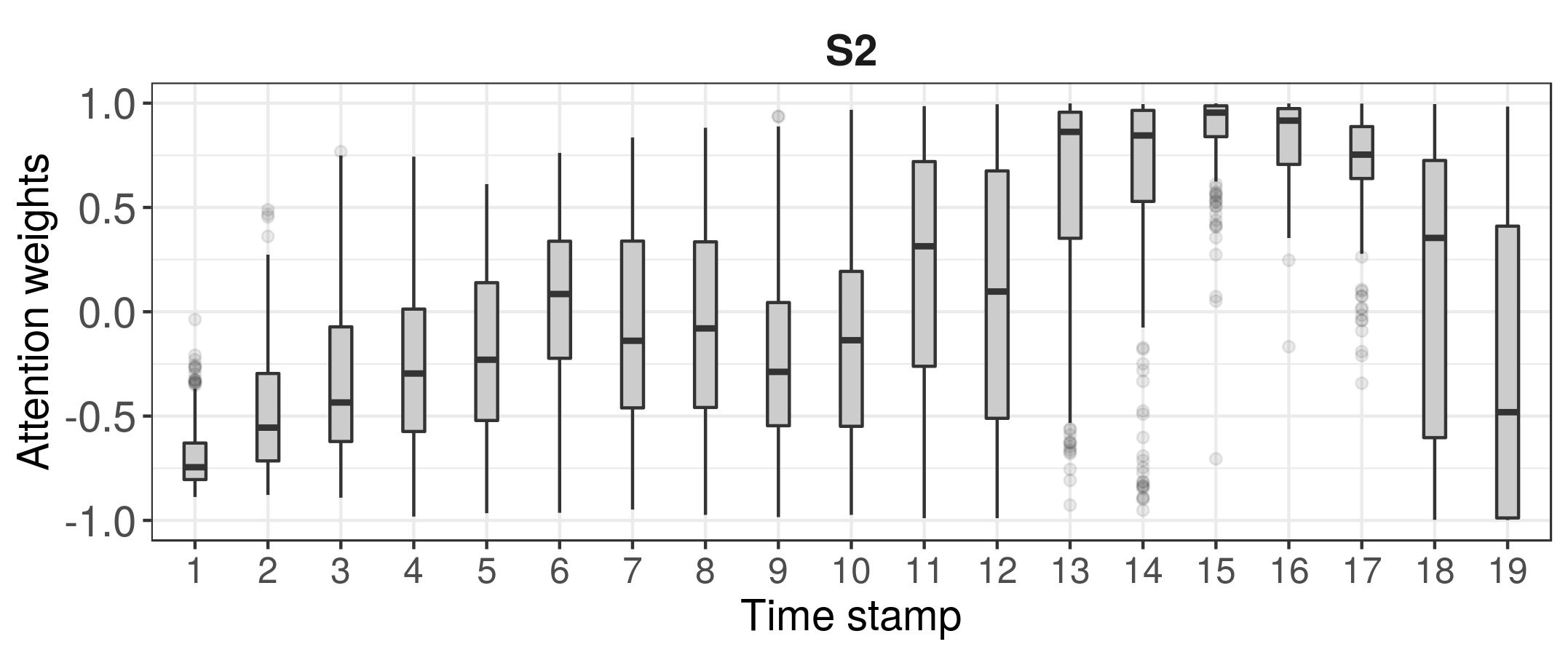}}
\hfill
\subfloat[Concatenated sources\label{fig:att-legumes3}]{\includegraphics[width=.9\linewidth]{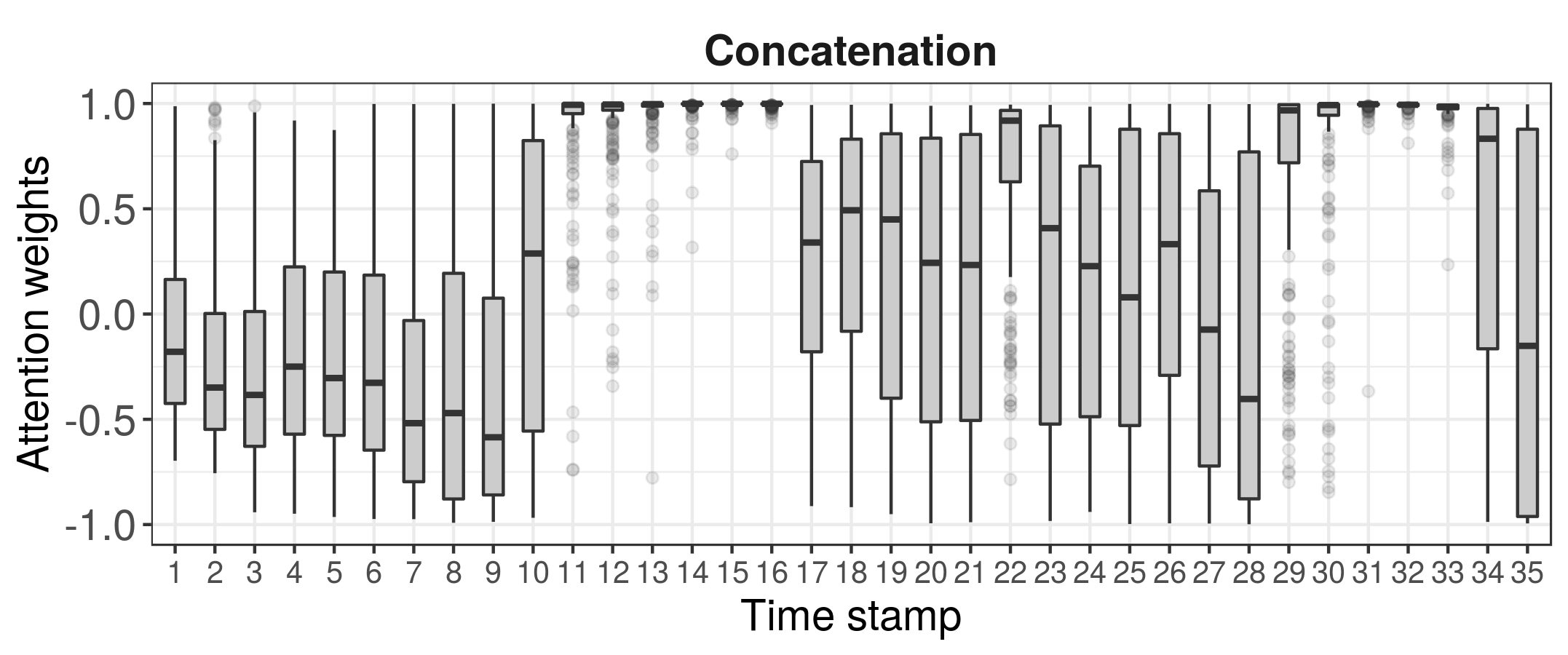}}
\caption{Box plots of the Legumes class attention weights for the radar (S1) SITS (16 time stamps), the optical (S2) SITS (19 time stamps) and the concatenated sources (radar followed by optical).}
\label{fig:att-legumes}
\end{figure}

In this last part of our experimental results, we explore the side information provided by the attention mechanism introduced in Section~\ref{sec:method-mam}. In order to get meaningful insights about how \method{} handles the multi-source time series for the land cover classification task. Attention weights have been successfully employed in the field of NLP~\cite{Bahdanau14,BritzGL17,ChoiCB18} to explain which parts of the input signal contribute to the final decision of the RNN models. With the aim to set up an analogous analysis in the remote sensing time series classification context, we consider attention weights on the \textit{Senegalese} site with a particular interest on crops (cereals and legumes) motivated by the agronomic knowledge we got from discussions with field experts. 
Recall that \method{} have learnt three sets of weights related to the attention mechanism employed to support the (auxiliary) radar, the (auxiliary) optical and the fused classifiers.
To inspect the behavior of such attention mechanisms, we depict in Figures~\ref{fig:att-cereals} and~\ref{fig:att-legumes}, the distribution of the three sets of attention weights on cereals and legumes land cover, respectively, via box plots visualizations. 

At first look, we can observe that the model weights quite similarly the radar and optical time stamps on both classes. \textcolor{black}{We can also notice that} some time stamps towards the end of each time series are highly weighted. This behavior is especially \textcolor{black}{evident} on the concatenated sources distribution \textcolor{black}{(Figures \ref{fig:att-cereals3} and \ref{fig:att-legumes3}), where high attention weights are observed from time stamp 10 (2018/08/18) for the radar SITS (ranging from 1 to 16) and time stamp 28 (2018/08/16) for the optical SITS (ranging from 17 to 35) which both correspond to the mid-August period. }
This is interesting to note the likely correlation existing between these high attention values and the crops growth, since in the \textit{Senegalese groundnut basin}, vegetation reaches its peak activity during the August month also characterized by heavy rain amount, all inducing more \textcolor{black}{differentiation} among land cover classes. 

\begin{figure}[!htbp]
\centering
\begin{tabular}{cc}
\subfloat[2018/10/05\label{fig:illu-att1}]{\includegraphics[width=.4\linewidth]{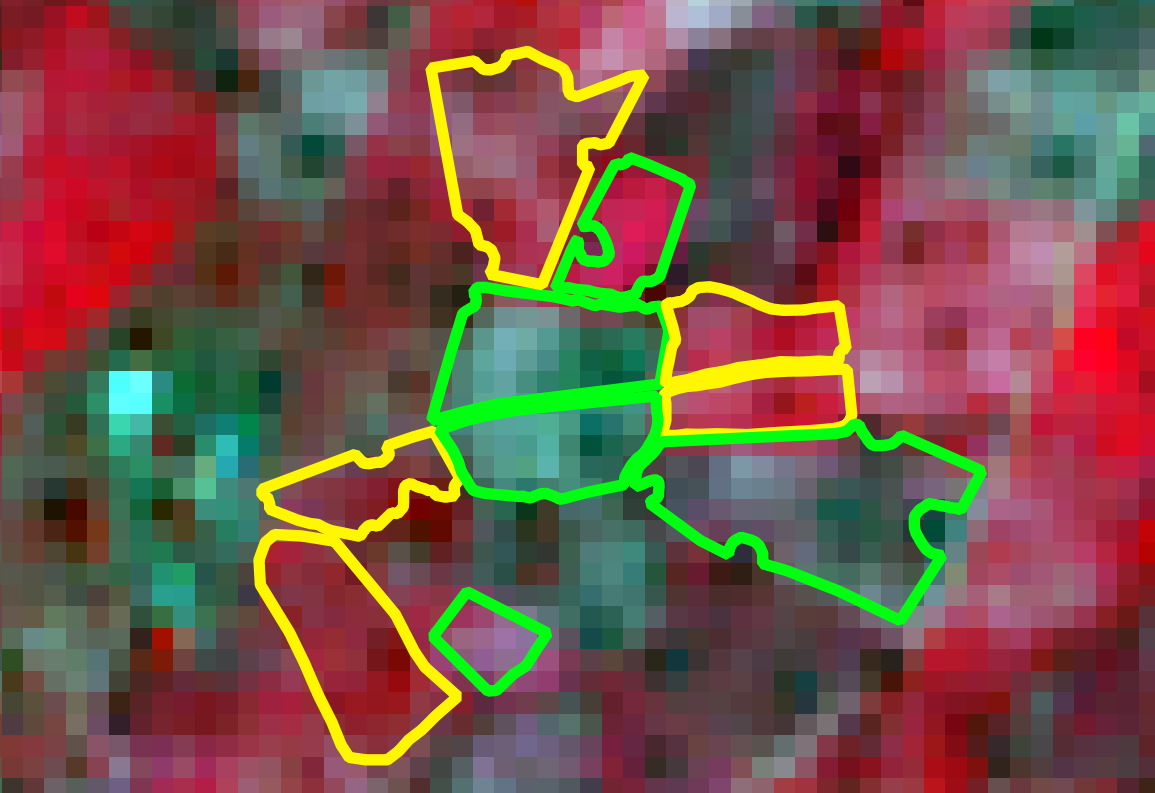}} 
&
\subfloat[2018/10/10\label{fig:illu-att2}]{\includegraphics[width=.4\linewidth]{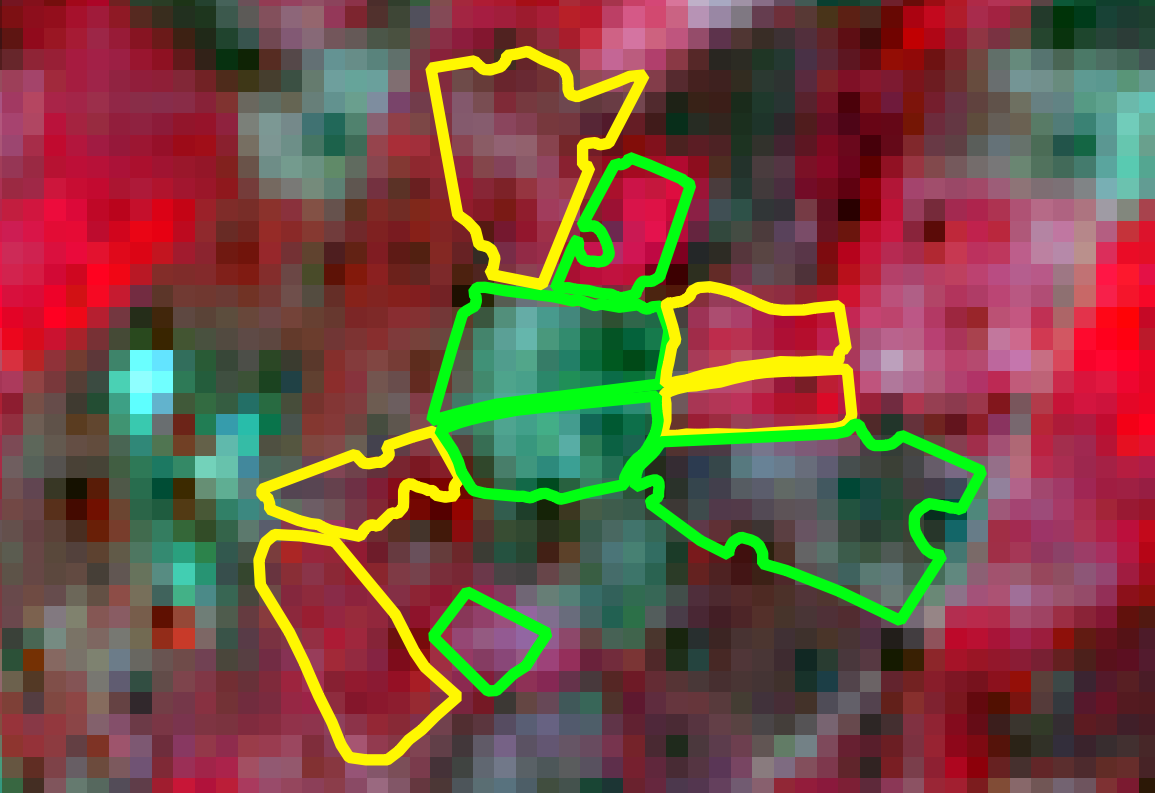}}
\\
\subfloat[2018/10/25\label{fig:illu-att3}]{\includegraphics[width=.4\linewidth]{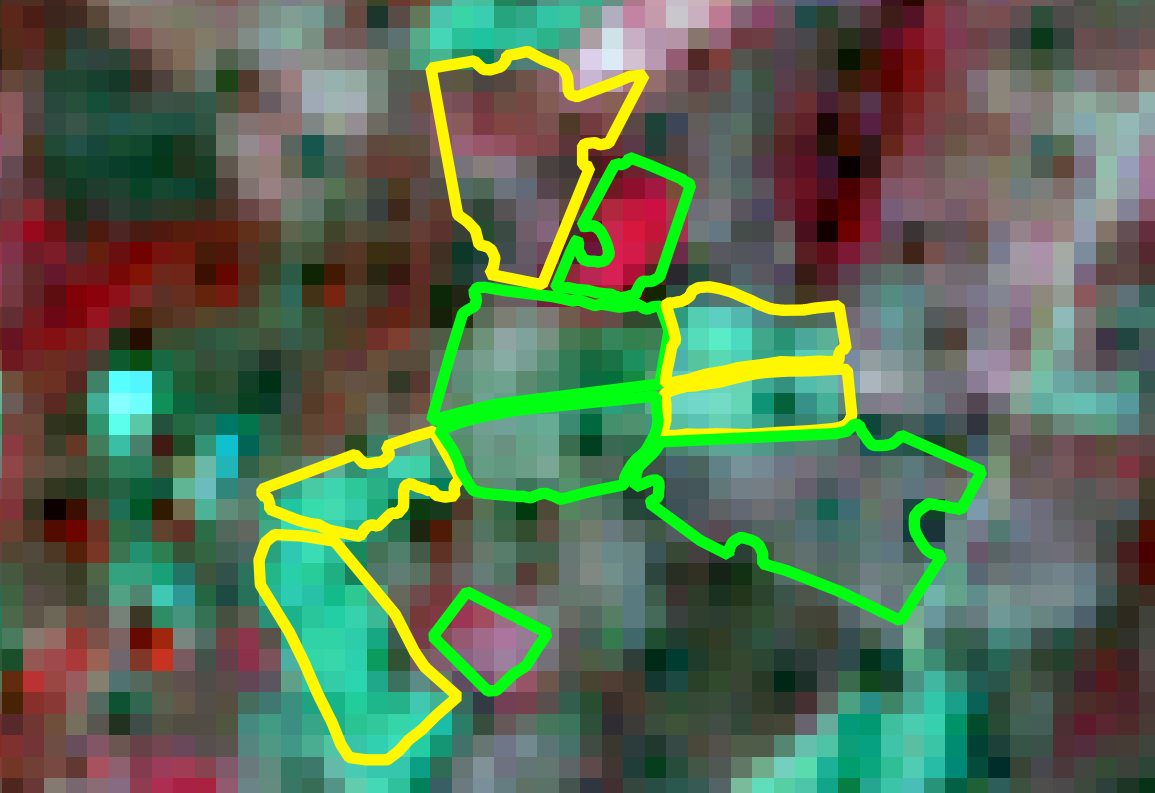}}
&
\subfloat[2018/10/30\label{fig:illu-att4}]{\includegraphics[width=.4\linewidth]{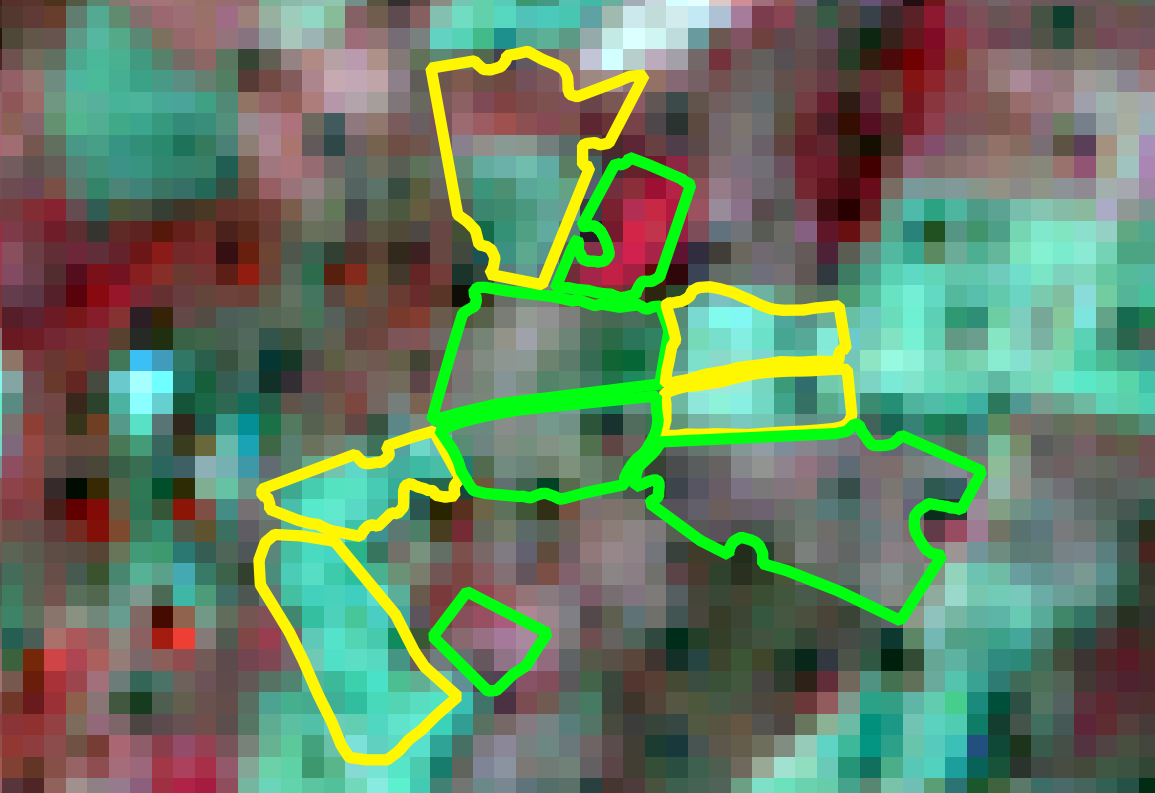}}
\\
\multicolumn{2}{c}{\subfloat[\label{fig:illu-att5}]{\includegraphics[width=.55\linewidth]{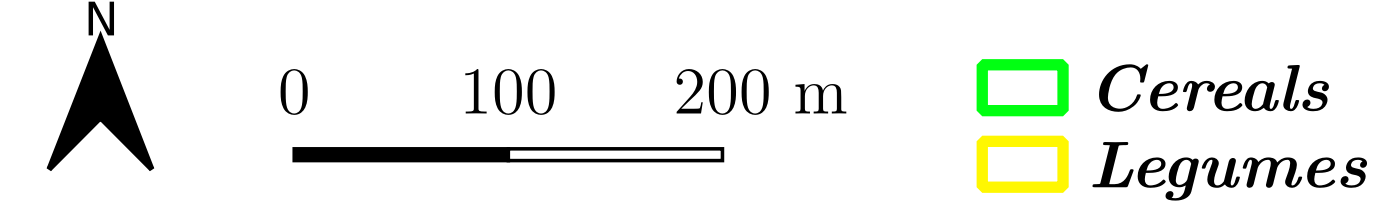}}}
\end{tabular}
\caption{Visualization of end of season agricultural practices in the \textit{Senegalese groundnut basin} concerning cereals and legumes. Background images come from the Sentinel-2 time series and are displayed in black--Red--Infrared composite colors. }
\label{fig:illu-att}
\end{figure}

However, on the two last optical time stamps (2018/10/25 and 2018/10/30), attention weights are \textcolor{black}{differently attributed} considering the 2 crop types. Cereals get more important attention weights for these two timestamps than legumes. This behavior seems to be \textcolor{black}{associated} to the agricultural practices adopted in the \textit{Senegalese groundnut basin} at the end of the season. While cereals (mainly millet) are harvested cutting only the ears, legumes (mainly groundnut) are torn off. Thus, on these latter time stamps, cereal plots are covered by senescent plants while legume plots turn into bare soil. These practices are visible in the SITS and illustrated in Figure~\ref{fig:illu-att}. 

To wrap up this analysis, we have found an existing correlation between the attention weights learnt by \method{} and the informativeness of the different time stamps in the SITS data. 
\textcolor{black}{To the best of our knowledge, no previous work exists on the exploitation of such kind of side information (attention weights) to understand and interpret the behavior of recurrent neural network models in the context of satellite image time series.}
As underlined by our findings, the exploration of the attention parameters can support the understanding of the decision made by our model and provides useful insights about the information contained in the SITS.

\subsection{Findings summary and evaluation discussion}

To summarize, the proposed deep learning framework
exhibits convincing performances in land cover mapping considering situation characterized by a realistic amount of available training samples. The comparison with other machine learning approaches underlines two points: i) our approach clearly outperforms the RF classifier that is the common approach employed to deal with SITS classification and ii) other standard, less explored, machine learning methods, i.e. SVM and MLP, exhibit competitive behaviors w.r.t. our method on the study site involving the small amount of labeled samples.

The ablation study indicates that \method{} is capable to exploit the complementarity between the radar and optical information always improving its performances w.r.t. using only one of the two sources. Our framework integrates background knowledge via hierarchical pretraining leveraging taxonomic relationships between land cover classes. Experiments highlight that such knowledge seems valuable for black box models and it systematically ameliorates the behavior of \method{}. On the other hand, some other type of considered knowledge, i.e. NDVI radiometric index, seems less effective due to the fact that, probably, the model is capable to autonomously derive it. These points clearly pave the way to further investigation about which and how knowledge can be injected to guide/regularize the learning process of such techniques.

Considering model interpretability, we also conduct some qualitative studies about the side information that can be extracted from our framework. The qualitative results we obtain are in line with the agronomic knowledge on the study area. Make the black box grey is an hot topic today in the machine learning community~\cite{Ribeiro0G16} and we can state, with a certain margin of confidence, that solutions or answers associated to this question will be available in a near future.

Finally, we remind that operational/realistic constraints might be considered when dealing with remote sensing analysis. Constraints can be related to available resources, i.e. timely production of land cover maps or limited access to training samples. We are aware that, in operational/realistic scenario characterized by the, almost real-time, production of land cover maps (i.e. disaster management~\cite{boccardo2015}), more computationally efficient solutions needs to be preferred (i.e. MLP or SVM) to deep learning approaches.
On the other hand, in our work we deal with (agricultural-oriented) land cover mapping where, land cover maps need to be provided with a relative low time frequency (once or twice per year). Due to this fact, here, the operational constraints are mainly intended regarding the limited amount of available labeled samples. In such data paucity setting, our approach clearly outperforms the Random Forest classifier, that is the de facto strategy involved in the classification of SITS data~\cite{BELGIU201624}. In addition, the experimental evaluation pointed out that, less explored machine learning techniques, in the context of SITS analysis, i.e. SVM and MLP, deserve much attention since they constitute valuable strategies to which compare future proposals. 

\section{Conclusion} \label{sec:concl}
In this work, we propose to deal with land cover mapping at object level, from multi-temporal and multi-source (radar and optical) data. Our approach is based on an extension of RNN involving a modified attention mechanism devised to better suit the SITS data context. We also introduce a novel hierarchical pretraining approach for neural networks which integrates specific knowledge from land cover classes to support the land cover mapping task. 
Extensive quantitative and qualitative evaluations on two different study sites demonstrate the effectiveness of the proposal compared to common and not common machine learning techniques in the field of land cover mapping.
As future work, we plan to investigate other deep learning approaches conceived to better deal with sequential (temporal) data than RNN namely transformers~\cite{VaswaniSPUJGKP17} or one dimensional convolutional neural networks~\cite{Pelletier2019} also tailored to process this kind of data. In addition, a possible extension of the actual framework could be done towards leveraging spatial dependencies in the multi-source SITS via convolutions.  

\section*{Acknowledgements}
This work was supported by the French National Research Agency under the Investments for the Future Program, referred as ANR-16-CONV-0004, the Programme National de Télédétection Spatiale (PNTS, http://www.insu.cnrs.fr/pnts), grant no PNTS-2018-5, the LYSA project (DAR-TOSCA) funded by the French Space Agency (CNES), the SERENA project funded by the Cirad-INRA metaprogramme GloFoodS and the SIMCo project (agreement number 201403286-10) funded by the Feed The Future Sustainable Innovation Lab (SIIL) through the USAID AID-OOA-L-14-00006. 
\bibliography{ref}

\end{document}